%% file: main.tex
\documentclass[10pt,twocolumn,letterpaper]{article}

\usepackage[pagenumbers]{wacv} 

\usepackage{booktabs}
\usepackage{threeparttable}
\usepackage{multirow}
\usepackage{graphicx}
\usepackage{microtype}
\usepackage[pagebackref,breaklinks,colorlinks,allcolors=wacvblue]{hyperref}
\usepackage{setspace}

\usepackage[accsupp]{axessibility}  


\definecolor{wacvblue}{rgb}{0.21,0.49,0.74}

\def\wacvPaperID{2153}
\def\confName{WACV}
\def\confYear{2027}

\title{RADIOlogy: A Single-Pass Approach to Training-Free Zero-Shot Anomaly Detection}

\author{Joaquín Figueira \quad Giacomo D'Amicantonio \quad Egor Bondarau \\
AIMS Group, Department of Electrical Engineering, Eindhoven University of Technology \\
{\tt\small \{j.figueira, g.d.amicantonio, e.bondarev\}@tue.nl}}

\begin{document}
\maketitle
\input{sec/0_abstract}    
\input{sec/1_intro}
\input{sec/2_previous_work}
\input{sec/3_method}
\input{sec/4_experiments}
\input{sec/5_conclusions}

{
    \small
    \bibliographystyle{ieeenat_fullname}
    \bibliography{main}
}

\clearpage
\setcounter{page}{1}
\renewcommand{\thepage}{S\arabic{page}}
\twocolumn[{%
\centering
{\Large \bf Supplementary Material for RADIOlogy: A Single-Pass Approach to\\ Training-Free Zero-Shot Anomaly Detection\par}
\vspace{2em}
}]
\setcounter{section}{0}
\setcounter{figure}{0}
\setcounter{table}{0}
\setcounter{equation}{0}
\renewcommand{\thesection}{S\arabic{section}}
\renewcommand{\thefigure}{S\arabic{figure}}
\renewcommand{\thetable}{S\arabic{table}}
\renewcommand{\theequation}{S\arabic{equation}}
\input{sec/6_supplementary}

\end{document}

%% file: sec/0_abstract.tex
\begin{abstract}
Current zero-shot anomaly detection methods often rely on complex prompt-learning techniques or computationally expensive sliding-window strategies to achieve fine-grained visual alignment to detect small-scale anomalies. This reliance stems from the fundamental limitation of standard CLIP architectures, where \texttt{[CLS]} token attention and spatial patch pooling struggle to capture subtle, localized anomaly features. While prompt-learning approaches aim to bridge this gap, they typically require a large number of ground-truth segmentation masks from auxiliary datasets, which are often unavailable in specialized domains. In this work, we show an emerging property of multi-head attention pooling in vision-language models: remarkable zero-shot anomaly localization and detection right out of the box. We translate this property into a single-pass anomaly detection technique, RADIOlogy, that does not require dense, windowed execution and significantly outperforms other training-free baselines. Instead, the method relies on a simple contrastive scoring approach that compares the average similarity between vision features and a suite of normal and abnormal prompts. Furthermore, we leverage the agglomerative features of RADIO models to enhance pixel-level anomaly segmentation using a dense anomaly refinement pipeline. Finally, we show how the vision-language family of training-free zero-shot anomaly detection techniques can be applied successfully to the medical domain with minimal modifications. Code is available at \url{https://github.com/jacomof/RADIOlogy}.
\end{abstract}

%% file: sec/1_intro.tex
\section{Introduction}
\label{sec:intro}

Visual Anomaly Detection (AD) aims to identify and localize irregular visual patterns in images, which are often associated with anomalies~\cite{bergmann2019mvtec}.
Because anomalous occurrences are inherently rare and structurally unpredictable, traditional approaches model the distribution of normal appearances via unsupervised or self-supervised representations~\cite{roth2022patchcore, damm2025anomalydino, defard2021padim, guo2025dinomaly, lendering2026subspacead}.
Recent breakthroughs in multi-modal foundation models and zero-shot transfer have allowed for a relatively new family of techniques to emerge: Zero-Shot Anomaly Detection (ZSAD)~\cite{jeong2023winclip, zhou2024anomalyclip}, methods that identify out-of-distribution defects without access to reference normal samples from the target distribution. Within ZSAD, existing paradigms broadly bifurcate into cross-dataset trained methods~\cite{cao2024adaclip, ma2025aa}, which learn auxiliary defect priors from separate annotated datasets, and training-free methods, which rely strictly on the intrinsic representations of frozen foundation models~\cite{jeong2023winclip, li2024fade, hicsonmez2026training}.

\begin{figure}
    \centering
    \includegraphics[width=1.0\linewidth]{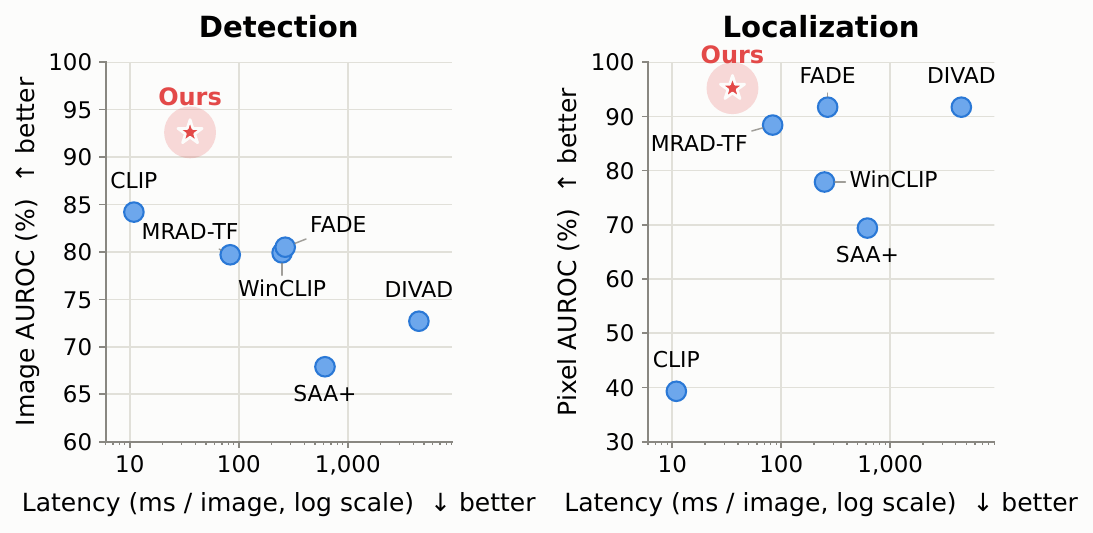}
    \caption{\textbf{Performance vs. latency comparison.} Anomaly segmentation and classification performance relative to detection latency for RADIOlogy (depicted with a star marker), five published training-free baselines and a naive CLIP approach on a seven-dataset industrial suite. Our method outperforms the baselines by a significant margin, while being at least two times faster than all methods except the naive CLIP approach.}
    \label{fig:teaser-auroc-throughput}
    \vspace{-5pt}
\end{figure}

Training-free approaches are particularly compelling for real-world deployments: they bypass the operational overhead and computational expense of auxiliary training~\cite{jeong2023winclip}, adapt immediately to severe domain shifts~\cite{xu2026mrad}, and eliminate the requirement for curated defect annotations in specialized downstream domains~\cite{cao2024adaclip}.
Until recently, training-free pipelines were anchored by WinCLIP~\cite{jeong2023winclip}, which queries the weakly supervised vision-language model (VLM) CLIP~\cite{radford2021learning} against predefined prompt ensembles.
However, a fundamental limitation of the standard CLIP architecture is its weak patch-level vision-language alignment~\cite{zhou2022extract,wysoczanska2024clipdenoiser}.
To circumvent the noisy localization of raw patch embeddings, WinCLIP relies on an expensive multi-scale sliding-window execution, extracting the \texttt{[CLS]} token over a large number of local crops per image.
While effective at bypassing CLIP's poor dense features, this multi-pass windowing introduces inference latency and computational overhead.
Consequently, recent works leveraging CLIP have shifted toward prompt-tuning or fine-tuning vision-language backbones on auxiliary anomaly corpora~\cite{zhou2024anomalyclip,ma2025aa}, largely abandoning training-free architectures.

In this work, we demonstrate that modern advances in vision-language architectures render complex windowing pipelines unnecessary, with the ultimate objective of reinvigorating language-aligned, training-free anomaly detection.
By investigating the Multi-Head Attention Pooling (MAP) mechanism adopted in the SigLIP family~\cite{zhai2023sigmoid,tschannen2025siglip} and consolidated within modern agglomerative foundation models, such as C-RADIOv4~\cite{ranzinger2026c}, we uncover an emergent capability: MAP-pooled encoders produce inherently well-aligned dense vision-language anomaly features directly from a single forward pass.
Because MAP attends to individual patch tokens via a learned summary query, each token retains a strong standalone semantic alignment with text prompts. This eliminates the structural bottleneck that hinders standard CLIP backbones.

Leveraging these properties, we present a unified, training-free zero-shot anomaly detection framework, RADIOlogy.
For image-level detection, our pipeline scores global anomaly margins in a single forward pass, setting a new state of the art among training-free methods across benchmark datasets without crop ensembles.
For pixel-level segmentation, we exploit the complementary teacher heads natively exposed by C-RADIOv4. Its raw trunk features provide content-adaptive affinity graphs to sharpen defect boundaries, while SAM3-aligned features enable object-part clustering to preserve structural boundaries.
Together with a training-free foreground prior that suppresses background noise, our approach bridges the performance gap between multi-crop and single-pass training-free anomaly detection methods.
Finally, we observe that prior evaluations of training-free VLM methods in medical domains have been artificially constrained by inadequate industrial priors.
We show that, in the context of our specific pipeline, substituting domain-appropriate clinical terminology into the prompt ensemble yields state-of-the-art medical anomaly detection in the training-free zero-shot regime without per-dataset prompt engineering.

To summarize, our primary contributions are:
\begin{itemize}
    \item We introduce a contrastive anomaly scoring strategy that relies on the difference in alignment between normal and abnormal prompts, rather than on absolute similarity scores.  
    \item We formulate an anomaly segmentation refinement strategy that exploits the multi-teacher representations of \mbox{C-RADIOv4}.
    \item We identify MAP-based vision-language encoders as superior backbones for zero-shot anomaly detection in a single-pass regime.
    \item We demonstrate that VLM ZSAD generalizes effectively to the medical domain in the training-free regime with minimal modifications.
\end{itemize}

%% file: sec/2_previous_work.tex
\section{Related Work}
\label{sec:related-work}

\paragraph{Visual Anomaly Detection.}
Visual Anomaly Detection (AD) encompasses various operational regimes defined by the availability of target-domain data~\cite{bergmann2019mvtec,roth2022patchcore}.
Under standard unsupervised AD, models have access to a clean set of nominal training samples to construct a normality representation against which the out-of-distribution deviations are identified~\cite{defard2021padim,zavrtanik2021draem}.
Batched zero-shot settings evaluate small, unlabeled target sets containing mixtures of normal and anomalous images~\cite{li2024musc}.
In contrast, pure Zero-Shot Anomaly Detection (ZSAD) operates under the strictest constraint: identifying and localizing anomalies in target datasets without access to any target-domain training samples whatsoever~\cite{jeong2023winclip}.

\paragraph{Zero-Shot Anomaly Detection Paradigms.}
Within ZSAD, two primary methodologies have emerged: auxiliary-trained methods and training-free methods. Auxiliary-trained approaches rely on cross-dataset generalization by learning defect representations on unrelated base datasets.
These include prompt-tuning techniques that optimize continuous text/visual prompts~\cite{zhou2024anomalyclip,cao2024adaclip}, full-network fine-tuning across multimodal components~\cite{ma2025aa,zhu2025fine}, and synthetic defect injection schemes that simulate surface anomalies~\cite{fuvcka2026anomalyvfm}.
While effective, auxiliary training incurs substantial adaptation costs and risks overfitting to synthetic or domain-specific defect archetypes.

Training-free methods, conversely, leverage the implicit world knowledge and semantic priors embedded within pretrained foundation models.
Foundational works, such as WinCLIP~\cite{jeong2023winclip} and FADE~\cite{li2024fade}, query frozen CLIP representations against predefined compositional text prompts.
Recent extensions have explored generative reconstruction via diffusion inversion (e.g., DIVAD~\cite{hicsonmez2026training}) or non-parametric patch memory retrieval (e.g., MRAD-TF~\cite{xu2026mrad}).
However, generative and sliding-window architectures remain computationally prohibitive during inference, while raw CLIP features exhibit severe spatial degradation when evaluated densely.
Our framework advances the training-free ZSAD paradigm by eliminating windowed inference while achieving state-of-the-art segmentation quality among training-free methods.

\paragraph{Vision-Language and Agglomerative Foundation Models.}
Vision-Language Models (VLMs) map visual and textual modalities into a shared metric embedding space via joint contrastive objectives~\cite{radford2021learning,jia2021scaling, fang2024dfnclip}.
Standard CLIP backbones extract global image summaries using a dedicated \texttt{[CLS]} token~\cite{radford2021learning, sun2023eva,fang2024dfnclip}.
While effective for holistic scene understanding, the unconstrained self-attention in standard ViTs causes localized patch embeddings to degrade in semantic specificity, hindering fine-grained patch-text alignment~\cite{zhou2022extract}.
To address these representation bottlenecks, the SigLIP architecture~\cite{zhai2023sigmoid,tschannen2025siglip} replaces the pairwise softmax contrastive loss with a pairwise sigmoid loss and adopts MAP.
MAP summarizes visual sequences via learned query probes, decoupling individual patch representations from global token mixing and, thereby, preserving localized semantic alignment~\cite{lee2019set,zhai2023sigmoid}.

In parallel, foundation model distillation has yielded agglomerative multi-modal encoders that unify distinct capabilities into a single monolithic backbone~\cite{ranzinger2024radio,wang2024samclip,sariyildiz2024unic,shang2024theia}.
Exemplified by the RADIO framework~\cite{ranzinger2024radio, heinrich2025radiov2, ranzinger2026c}, these architectures employ lightweight projection adapters to simultaneously emulate diverse teacher spaces, including zero-shot segmentation (SAM~\cite{kirillov2023segment,carion2026sam}), self-supervised dense geometry (DINOv2/v3~\cite{oquab2023dinov2, simeoni2025dinov3}), and open-vocabulary language alignment (SigLIP2~\cite{tschannen2025siglip}).
By leveraging the unified feature space of C-RADIOv4, our work demonstrates that the synergy between MAP language alignment, the vision backbone's structural affinities, and SAM part segmentation enables comprehensive, single-pass zero-shot anomaly localization without specialized network fine-tuning.

%% file: sec/3_method.tex
\section{Method}
\label{sec:method}

\begin{figure*}[t]
  \centering
  \includegraphics[width=\linewidth]{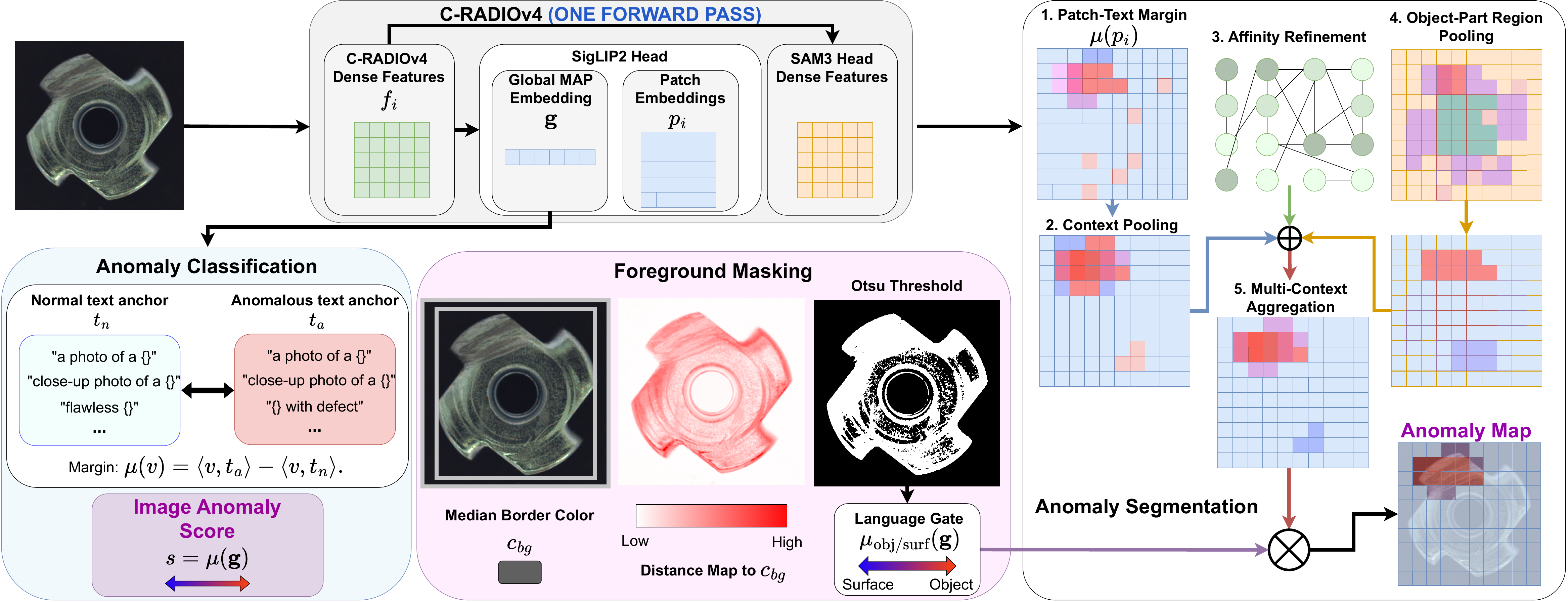}
  \caption{\textbf{Overview of our proposed method.} First, the input image is encoded by the agglomerative encoder and two of the three teacher heads (SigLIP2 and SAM3). Using the language-aligned features and the prompt ensemble a margin score is computed per token-embedding to obtain the image-level anomaly score and the raw patch-level margin scores (1). Then, pixel-level refinement is applied to the patch scores using context pooling (2), affinity refinement (3) and finally object-part pooling (4). The resulting scores are aggregated per patch and the final anomaly mask is obtained by applying the color-based foreground segmentation mask to the patch scores.} 
  \label{fig:method}
  \vspace{-5pt}
\end{figure*}

Our pipeline, depicted in \cref{fig:method}, produces both an image-level anomaly score and a pixel-level anomaly
map from a single forward pass through an agglomerative encoder. Section~\ref{sec:method-prelim} fixes notation
and the prompt ensemble, which we inherit from WinCLIP~\cite{jeong2023winclip}. Section~\ref{sec:method-image} describes how the MAP embeddings are used as an 
anomaly classifier. Section~\ref{sec:method-pixel} builds the anomaly map from
the same forward pass by combining patch-text scoring with two auxiliary signals
 from C-RADIOv4~\cite{ranzinger2026c}. These include the raw dense 
features of C-RADIOv4 for spatial refinement and its SAM3-aligned features for object-part
clustering. Section~\ref{sec:method-mask} adds a training-free foreground prior
that suppresses the background after scoring. Section~\ref{sec:method-medical}
describes the single substitution that ports the whole pipeline to the medical
domain.

\subsection{Preliminaries}
\label{sec:method-prelim}

\paragraph{Text Anchors.}
Given a test image $x$, a SigLIP2-compatible vision backbone $\phi$ produces a set of $N$ patch embeddings
$\{p_i\}_{i=1}^N \subset \mathbb{R}^d$, arranged on a $g_h \times g_w$ grid, and a
single global embedding $\mathbf{g} \in \mathbb{R}^d$, both already projected into
the joint vision-language space of SigLIP2~\cite{tschannen2025siglip}. For
the main encoder employed, C-RADIOv4, $\phi$ is the combination of the distilled trunk and the SigLIP2-aligned projection head.

Following WinCLIP, we build a text ensemble from 22 photographic
templates (\eg ``a photo of a \{\}'', ``a close-up photo of the \{\}'') crossed
with a small set of normal and anomalous state phrases (\eg ``flawless \{\}'',
``\{\} with defect''), with the category name substituted into \{\}. For each
category $c$, every template-state combination is encoded, 
the embeddings are $\ell_2$-normalized, and the mean of normal and abnormal prompts is computed. This initial $\ell_2$-normalization is performed before averaging, so that large prompts do not dominate the ensemble. This produces two unit vectors per category, $t_n$ and
$t_a$, the normal and anomalous text anchors. We reuse WinCLIP's original
industrial vocabulary verbatim; the only substitution we incorporate is a domain-wide
substitution described in Section~\ref{sec:method-medical}.

For a unit-norm vector $v$ obtained from a vision embedding, we define the anomaly margin
\begin{equation}
    \mu(v) = \langle v, t_a \rangle - \langle v, t_n \rangle .
    \label{eq:margin}
\end{equation}
As the two text anchors share all common elements describing the category, \eg
object identity, their difference cancels that shared component and results in the
axis that separates normal from anomalous appearance. We use $\mu$ for both the
image-level score and every step of the pixel-level map with different inputs.

\paragraph{Dense Alignment of the MAP Head.}
We leverage the dense features of a MAP head-aligned vision encoder, obtained by a single pass through the model, to produce our anomaly detection mechanism. A MAP head summarizes a grid of tokens with a learned query that attends to each patch token produced by the vision backbone. The output of this head is a global representation of an image that is aligned with textual representations, using either a binary classification or a contrastive loss. In its simplest form, the pre-training task consists of classifying whether an input image and input text are related. Given that the token embeddings function as direct keys and values for the MAP head, and the head is relatively shallow compared to the vision backbone, it is forced to align the patch token representations aggressively to obtain a high-quality summary representation. This is structurally different from CLIP's original recipe~\cite{radford2021learning},
where the \texttt{[CLS]} summary token is mixed with every patch through
repeated self-attention blocks. The result is a more diluted relationship between the \texttt{[CLS]} token and each patch token, culminating in weak dense alignment. Note that although C-RADIOv4 does not use a MAP head internally, its alignment with the MAP module of SigLIP2 preserves its dense feature quality.

\subsection{Anomaly Classification}
\label{sec:method-image}

We score an image with a single number, $s = \mu(\mathbf{g})$, using the MAP-pooled
global embedding directly against the text anchors. This bypasses the standard practice in VLM-based anomaly detection pipelines, which employ a combination of dense and global features to obtain anomaly classification scores~\cite{jeong2023winclip, li2024fade}.

\subsection{Anomaly Segmentation}
\label{sec:method-pixel}

We leverage the strong, dense features produced by the MAP-aligned vision encoder to obtain an anomaly score for each patch using a single forward pass. 

The image-level score described in \cref{sec:method-image} discards spatial information by construction. To
localize anomalies, we score patches directly using the same margin formulation and refine the results with two
signals produced by the same encoder pass: C-RADIOv4 raw 
dense features and SAM3-aligned dense features. For a patch embedding $p_i$ produced by the vision backbone, we define its raw C-RADIOv4 features as $f_i$.

\paragraph{Patch-text Margin and Context Pooling.} The raw pixel map applies~\cref{eq:margin} to every patch embedding $p_i$,  $m_i = \mu(p_i)$, where $m_i$ denotes the margin score of $p_i$. On its own, this map is
semantically informative but spatially noisy: a single patch does not always contain sufficient semantic information to represent an anomalous concept. We recover context by average-pooling
patch embeddings over a $k \times k$ neighborhood $W_i$ before scoring:
\begin{equation}
    \bar{p}_i = \frac{1}{|W_i|} \sum_{j \in W_i} p_j, \qquad m_i^{k} = \mu(\bar{p}_i).
    \label{eq:context}
\end{equation}

\paragraph{Affinity refinement.} A square window mixes objects and backgrounds
indiscriminately, resulting in low-resolution defect boundaries.
To tackle this, we replace the square window with a content-adaptive one, built from RADIO's
dense raw features $\{f_i\}$, which are better suited for identifying semantically related visual concepts. We build a patch affinity matrix:
\begin{equation}
    A_{ij} = \frac{\exp(\hat{f}_i \cdot \hat{f}_j / \tau)}{\sum_{k \neq i} \exp(\hat{f}_i \cdot \hat{f}_k / \tau)}, \qquad j \neq i,
    \label{eq:affinity}
\end{equation}
where $\hat f$ denotes $\ell_2$-normalized features, $\tau$ is a temperature parameter determining the uniformity of the affinity distribution. The diagonal of $A$ is excluded
so that a patch's identical self-similarity cannot collapse the softmax onto the
identity. We then diffuse the margin field $m$ over this graph:
\begin{equation}
    D : m \leftarrow \alpha \, m + (1-\alpha) \, A m,
    \label{eq:diffuse}
\end{equation}
where $\alpha$ is a decay coefficient determining the strength of the diffusion process $D$. This process, inspired by CLIP-DINOiser's~\cite{wysoczanska2024clipdenoiser} semantic segmentation refinements, is performed 
for a small fixed number of iterations, such that the information embedded in a patch token is enriched by the information from its neighboring patches. This sharpens a
defect to its true extent by allowing the sharing of information between semantically related tokens.

\paragraph{Object-part Region Pooling.} Context pooling and affinity refinement operate on fixed neighborhoods and ignore the boundaries of individual objects. To address this, we leverage spherical $k$-means over RADIO's SAM3-aligned features. As SAM3 was trained for segmentation, its features are better suited for this task. With this clustering algorithm, the image's patch grid is partitioned into $K$ clusters that respect object structure rather than a fixed or diffused region. 
We pool the
text-aligned embeddings within each cluster before scoring:
\begin{equation}
    z_i^K = \mu \!\left( \frac{\sum_{j: c(j) = c(i)} \mathbf{p}_j}{\left\| \sum_{j: c(j) = c(i)} \mathbf{p}_j \right\|_2} \right),
    \label{eq:region}
\end{equation}
where $c(i)$ is patch $i$'s cluster assignment, so a whole part is scored as a
unit. A piecewise-constant map discards the gradient within a region; therefore, we 
average the region score with the $k{=}5$ context field from
Equation~\ref{eq:context}:
\begin{equation}
    m_i^{K,\text{mix}} = \tfrac{1}{2}\left( z_i^K + m_i^5 \right).
    \label{eq:regionmix}
\end{equation}
To maximize the coverage over a wide configuration of object-part relationships, we then average this quantity over one or more cluster counts and
the affinity diffusion $D(\cdot)$ of Equation~\ref{eq:diffuse}. This is similar to WinCLIP's approach of  averaging across multiple window
scales, but without the computational overhead of multiple backbone forward passes per image. The final pre-masking patch-level scores are computed using the following four-way average,
\begin{equation}
    m_i^{\text{final}} = \frac{1}{4} \sum_{K \in \{8, 32\}} \left[ m_i^{K,\text{mix}} + D\!\left( m_i^{K,\text{mix}} \right) \right].
    \label{eq:finalmap}
\end{equation}

\subsection{Foreground Masking}
\label{sec:method-mask}

Given that image backgrounds are often featureless and lack distinctive semantics, applying a semantically-aligned patch-text margin to background patches results in anomaly score noise. To mitigate the effect of such distortion, we apply a foreground mask after applying the previously described patch scoring recipe. To this end, we propose a simple approach: a foreground prior computed exclusively from the image, with no backbone and no learned features.

For a test image, we take the median RGB color of a fixed-width band along the
image border and compute, at pixel resolution, the distance of every pixel from
that color. Otsu's~\cite{otsu1979threshold} method thresholds this field automatically, and a patch is
considered foreground if any of its pixels are above the threshold.

In an image with no background (\eg surface categories on MVTec-AD), Otsu's method becomes uninformative and often results in an arbitrary partition of the input image based on superficial characteristics.   Rather than introducing a separate detector for this, we reuse the mechanism from
Section~\ref{sec:method-image}. We score the same MAP-pooled global embedding $\mathbf{g}$ with the margin of~\cref{eq:margin}, substituting a second
text-anchor pair. This second pair, $t_{\text{obj}}$ and $t_{\text{surf}}$, is built from object-like words (\ie
``object'', ``thing'') and surface-like words (\ie
``surface'', ``texture''), respectively. More precisely, this language-guided background gate, defined over the MAP-pooled summary token $\mathbf{g}$, is: $\gamma = \mu_{\text{obj/surf}}(\mathbf{g})$.  Masking is disabled whenever $\gamma$ falls below a fixed threshold. As in the state vocabulary itself, this object/surface pair is declared a priori and swapped
once per domain. The only added cost is a 
text-encoder query for two anchors; no additional image forward pass is required.
Patches are marked as foreground if any of their pixels are within the foreground mask, and background patches have their margin replaced by the image's own
minimum score before upsampling. This is performed to ensure that a masked region can never outrank a genuinely normal foreground patch.

\subsection{Domain Vocabulary Adaptation}
\label{sec:method-medical}

Every step above depends on the text anchors $t_n, t_a$ exclusively through the state
vocabulary; the images, the encoder, and the aggregation are
maintained fixed for all domains. WinCLIP's state words presume a manufacturing setting, using descriptors such as ``damaged'' or ``with flaw'', which do not translate semantically to a medical setting. To accommodate the appropriate priors for the medical domain, we replace the industrial state words with a clinical set derived from domain knowledge (\eg ``\{\} with a lesion'',
``diseased \{\}'') and fix every other component of the pipeline.

%% file: sec/4_experiments.tex
\section{Experiments}
\label{sec:experiments}
\subsection{Experimental Setup}
\label{sec:exp-setup}

\begin{table*}[ht]
\centering
\caption{\textbf{Training-free zero-shot anomaly \emph{classification}, industrial and medical
benchmarks.} The table reports image-level AUROC~/~max-F1~/~AP (\%) for all anomaly detection benchmarks under six baseline techniques and our own method.  WinCLIP and SAA+ are each run under their own industrial vocabulary in the medical evaluations, which showed comparable or better overall performance than the medical vocabulary (see the supplementary material, \cref{sec:medical-vocabulary}), while our method employs the medical prompts. Bold marks the best training-free result per row; underline marks the
second-best. FADE~\cite{li2024fade} is excluded from the medical evaluations as explained in \cref{sec:exp-setup}.}
\label{tab:trainfree-image}
\resizebox{0.9\textwidth}{!}{%
\begin{tabular}{l ccccccc}
\toprule
Dataset & CLIP & SAA+ & WinCLIP & FADE & DIVAD & MRAD-TF & \textbf{Ours} \\
\midrule
\multicolumn{8}{l}{\textit{Industrial}} \\
\midrule
MVTec-AD      & 88.2 / 90.1 / 94.4 & 62.8 / 87.0 / 81.3 & \underline{90.4} / \underline{92.7} / \underline{95.6} & 90.1 / 92.0 / 95.4 & 75.9 / 87.3 / 86.2 & 79.0 / 88.2 / 90.0 & \textbf{96.0} / \textbf{95.1} / \textbf{98.0} \\
VisA          & \underline{79.7} / \underline{78.7} / \underline{83.6} & 70.9 / 76.0 / 77.2 & 75.5 / 78.2 / 78.7 & 74.7 / 78.3 / 78.3 & 69.7 / 74.2 / 75.4 & 75.0 / 77.6 / 79.1 & \textbf{85.4} / \textbf{83.1} / \textbf{87.8} \\
BTAD          & 75.2 / 72.9 / 74.8 & 43.5 / 64.4 / 53.0 & 68.2 / 67.8 / 70.9 & 74.3 / 69.6 / 71.7 & 82.4 / \underline{88.0} / \underline{91.3} & \underline{86.3} / 85.2 / 90.3 & \textbf{94.4} / \textbf{90.1} / \textbf{93.7} \\
MPDD          & \underline{73.1} / \underline{78.6} / \textbf{80.6} & 47.0 / 73.8 / 61.0 & 61.5 / 77.5 / 69.2 & 62.8 / 78.0 / 69.4 & 63.4 / 75.0 / 66.2 & 67.9 / 77.8 / 73.8 & \textbf{77.0} / \textbf{82.0} / \underline{78.9} \\
KSDD          & 83.0 / 54.2 / 54.9 & 68.0 / 31.4 / 25.1 & 77.0 / 41.2 / 35.2 & 75.4 / 38.9 / 30.8 & 67.1 / 34.3 / 20.2 & \underline{92.5} / \underline{68.1} / \underline{72.6} & \textbf{97.5} / \textbf{90.0} / \textbf{92.9} \\
DAGM          & \underline{95.3} / \underline{82.1} / \underline{85.1} & 89.7 / 80.5 / 80.0 & 91.8 / 75.7 / 79.5 & 90.9 / 73.5 / 77.4 & 71.2 / 40.9 / 31.5 & 78.4 / 57.6 / 57.8 & \textbf{99.6} / \textbf{95.8} / \textbf{98.0} \\
DTD-Synthetic & 95.1 / \underline{94.7} / \underline{98.2} & 93.6 / 93.0 / 95.9 & 95.1 / 94.1 / 97.7 & \underline{95.2} / 94.6 / 97.7 & 79.4 / 90.4 / 92.5 & 79.1 / 87.5 / 89.6 & \textbf{98.6} / \textbf{97.6} / \textbf{99.5} \\
\emph{Mean}   & \underline{84.2} / \underline{78.7} / \underline{81.7} & 67.9 / 72.3 / 67.6 & 79.9 / 75.3 / 75.3 & 80.5 / 75.0 / 74.4 & 72.7 / 70.0 / 66.2 & 79.7 / 77.4 / 79.0 & \textbf{92.6} / \textbf{90.5} / \textbf{92.7} \\
\midrule
\multicolumn{8}{l}{\textit{Medical}} \\
\midrule
HeadCT        & \underline{86.5} / \underline{82.9} / \underline{84.6} & 45.1 / 67.3 / 44.4 & 83.7 / 78.8 / 81.6 & -- & 65.7 / 69.9 / 66.4 & 69.3 / 67.5 / 73.6 & \textbf{98.0} / \textbf{93.9} / \textbf{98.4} \\
BrainMRI      & 94.0 / \underline{93.2} / 95.8 & 51.7 / 77.8 / 58.6 & 88.9 / 86.4 / 92.9 & -- & 75.3 / 82.4 / 79.0 & \underline{94.1} / 89.7 / \underline{96.8} & \textbf{96.0} / \textbf{94.9} / \textbf{97.4} \\
Br35H         & \underline{95.4} / \underline{90.2} / \underline{94.8} & 54.7 / 68.4 / 49.1 & 83.9 / 77.1 / 84.8 & -- & 71.6 / 70.9 / 69.2 & 91.1 / 82.3 / 92.2 & \textbf{97.9} / \textbf{94.7} / \textbf{97.9} \\
\emph{Mean}   & \underline{91.9} / \underline{88.7} / \underline{91.8} & 50.5 / 71.2 / 50.7 & 85.5 / 80.8 / 86.4 & -- & 70.9 / 74.4 / 71.5 & 84.8 / 79.8 / 87.5 & \textbf{97.3} / \textbf{94.5} / \textbf{97.9} \\
\bottomrule
\end{tabular}
}

\end{table*}

\paragraph{Datasets.}
We evaluate the seven industrial and seven medical benchmarks selected by  AdaCLIP~\cite{cao2024adaclip}, which provides a more comprehensive set than the standard training-free baselines~\cite{jeong2023winclip, li2024fade}. The industrial datasets consist of: MVTec-AD~\cite{bergmann2019mvtec}, VisA~\cite{zou2022spot},
BTAD~\cite{mishra2021vt}, MPDD~\cite{jezek2021deep}, KolektorSDD~\cite{tabernik2020segmentation} (KSDD), DAGM~\cite{wieler2007weakly}, and DTD-Synthetic~\cite{cimpoi2014describing, aota2023zero}. For the
medical domain, we adopt 
seven benchmarks, partitioned by task: HeadCT~\cite{campos2018headct}, BrainMRI~\cite{kanade2015brain}, and Br35H~\cite{hamada2020br35h} supply image-level labels only and are used for
detection; ISIC~\cite{gutman2016isic}, ClinicDB~\cite{bernal2015wm}, ColonDB~\cite{tajbakhsh2015automated}, and TN3K~\cite{gong2023tn3k} are entirely
anomalous with pixel masks and are used for localization. This is reflected in the medical block of \cref{tab:trainfree-image} and \cref{tab:trainfree-pixel}, which exhibit disjoint dataset subsets.

\paragraph{Aspect Ratio Correction.} During experimentation, we noticed that RADIOlogy's pipeline is sensitive to the very extreme aspect ratio resolution deviations present in the KSDD dataset (see \cref{sec:aspect-ratio-correction-ablation} in the  supplementary material). As a result, the input is split into square bands whenever the source aspect ratio departs
noticeably from 1:1. We apply one global, model-free rule: a frame whose native aspect ratio exceeds 2:1 is split into
$\lceil \text{long}/\text{short} \rceil$ non-overlapping bands along its long axis, each
scored independently and re-assembled for evaluation. 

\paragraph{Baselines.} All compared methods are training-free at inference, matching the scope of this work. CLIP is a deliberately simple ablation: a ViT-L/14@336 CLIP scored with our prompt ensemble and margin scoring, isolating the contribution of the backbone. WinCLIP~\cite{jeong2023winclip} is reproduced with AnomalyCLIP's~\cite{zhou2024anomalyclip} Accurate-WinCLIP re-implementation. SAA+~\cite{cao2023segment} is reproduced from the maintained
\texttt{SAA-plus} codebase unmodified. FADE and DIVAD are run end-to-end using the authors' released code with their reported hyper-parameters; since FADE's prompts extend WinCLIP's industrial vocabulary, we report it on the industrial block only rather than improvise a
medical adaptation. MRAD-TF uses a score fusion weight $k{=}0.5$, except on
VisA, where the code default $k{=}0.7$ reproduces the published result. Every method is evaluated on full-frame KSDD and DAGM's official test split. On the benchmarks, splits, and metrics each method's authors report, and for methods run at their published configuration, our reproductions are within 1 point of the published results, with two exceptions: DIVAD on BTAD, where our reproduction performs substantially better than the published numbers, and SAA+, where the reproduction performs slightly worse on MVTec-AD and considerably better on VisA (see the supplementary material, \cref{sec:baseline-reproduction}).

\paragraph{Metrics.} We report image-level Area Under the Receiver Operating Characteristic curve (AUROC), max-F1, and Average Precision (AP) for detection, as
well as pixel-level AUROC, max-F1, and Area Under the Per-Region Overlap curve (AUPRO) for localization, macro-averaged over each benchmark's
categories and then over all benchmarks. All three pixel metrics for RADIOlogy come
from a single shared-sort GPU implementation that matches \texttt{scikit-learn} to the sixth decimal place; baselines are scored using each method's released evaluation code.
\begin{table*}[t]
\centering
\caption{\textbf{Training-free zero-shot anomaly \emph{segmentation}, industrial and
medical benchmarks.} The table contains pixel-level AUROC~/~max-F1~/~AUPRO (\%) for a collection of anomaly segmentation benchmarks under our method and six training-free ZSAD techniques. WinCLIP and SAA+ are each run under their own industrial vocabulary in the medical evaluations, which showed comparable or better overall performance than the medical vocabulary (see \cref{sec:medical-vocabulary} of the supplementary material), while our method uses the medical prompts. FADE~\cite{li2024fade} is not included in the medical evaluations as explained in \cref{sec:exp-setup}.
Bold marks the best training-free result per row; underline marks the second-best.}
\label{tab:trainfree-pixel}
\resizebox{0.9\textwidth}{!}{%
\begin{tabular}{l ccccccc}
\toprule
Dataset & CLIP & SAA+ & WinCLIP & FADE & DIVAD & MRAD-TF & \textbf{Ours} \\
\midrule
\multicolumn{8}{l}{\textit{Industrial}} \\
\midrule
MVTec-AD      & 35.0 / 6.4 / 7.6 & 73.2 / 37.7 / 53.5 & 82.3 / 24.8 / 61.9 & \underline{89.2} / \textbf{39.8} / \textbf{84.1} & 88.1 / 35.8 / 72.4 & 86.7 / 28.1 / 63.5 & \textbf{92.2} / \underline{39.0} / \underline{82.9} \\
VisA          & 43.9 / 3.5 / 13.0 & 76.0 / \textbf{33.0} / 55.4 & 73.2 / 9.0 / 51.0 & 91.1 / 16.8 / \underline{79.1} & \underline{93.4} / \underline{24.0} / 78.2 & 91.0 / 15.7 / 71.0 & \textbf{93.7} / 20.4 / \textbf{83.1} \\
BTAD          & 30.3 / 6.2 / 11.8 & 58.4 / 8.1 / 26.4 & 72.7 / 18.5 / 27.5 & \textbf{96.3} / 46.4 / \underline{76.7} & 94.3 / \underline{52.5} / 69.9 & 80.5 / 15.7 / 36.3 & \underline{96.2} / \textbf{52.9} / \textbf{77.8} \\
MPDD          & 43.6 / 5.0 / 16.9 & 76.7 / 15.6 / 55.0 & 71.2 / 15.4 / 40.5 & \textbf{95.3} / 24.6 / \underline{83.9} & \underline{94.9} / \textbf{27.0} / 82.0 & 92.7 / 13.9 / 76.2 & 94.4 / \underline{24.7} / \textbf{85.5} \\
KSDD          & 71.1 / 1.5 / 31.4 & 68.3 / 6.2 / 30.5 & 79.1 / 7.0 / 34.3 & 84.4 / 6.4 / 56.1 & 88.2 / 4.5 / \underline{72.8} & \underline{90.6} / \underline{19.1} / 48.7 & \textbf{98.5} / \textbf{38.5} / \textbf{93.8} \\
DAGM          & 20.0 / 0.6 / 3.5 & 57.0 / \underline{29.4} / 34.0 & 87.6 / 12.7 / 65.7 & \underline{90.2} / 26.0 / \underline{79.2} & 88.8 / 15.2 / 74.5 & 86.7 / 23.6 / 64.9 & \textbf{95.3} / \textbf{43.7} / \textbf{88.6} \\
DTD-Synthetic & 31.2 / 4.5 / 12.3 & 75.9 / \textbf{59.3} / 57.1 & 79.5 / 16.1 / 51.4 & \underline{95.4} / \underline{50.6} / \underline{88.4} & 94.2 / 48.5 / 82.9 & 90.7 / 44.5 / 73.8 & \textbf{96.4} / 45.1 / \textbf{89.3} \\
\emph{Mean}   & 39.3 / 4.0 / 13.8 & 69.4 / 27.0 / 44.5 & 77.9 / 14.8 / 47.5 & \underline{91.7} / \underline{30.1} / \underline{78.2} & \underline{91.7} / 29.6 / 76.1 & 88.4 / 22.9 / 62.1 & \textbf{95.2} / \textbf{37.8} / \textbf{85.9} \\
\midrule
\multicolumn{8}{l}{\textit{Medical}} \\
\midrule
ISIC          & 46.7 / 44.0 / 12.4 & 78.0 / \underline{69.3} / 65.9 & 74.4 / 55.1 / 49.2 & -- & 71.5 / 53.7 / 40.0 & \underline{84.4} / 64.3 / \underline{72.1} & \textbf{90.5} / \textbf{74.0} / \textbf{76.9} \\
ClinicDB      & 64.3 / 22.9 / 29.0 & 78.7 / \underline{47.2} / 54.8 & 69.7 / 26.7 / 32.2 & -- & 74.3 / 28.8 / 44.5 & \underline{81.6} / 37.9 / \underline{59.0} & \textbf{89.6} / \textbf{54.9} / \textbf{73.2} \\
ColonDB       & 65.4 / 19.8 / 27.2 & \underline{78.2} / \underline{42.2} / 52.7 & 64.7 / 21.1 / 29.5 & -- & 71.1 / 23.0 / 52.4 & 77.6 / 30.4 / \underline{59.6} & \textbf{89.5} / \textbf{47.8} / \textbf{73.5} \\
TN3K          & 43.9 / 22.8 / 12.5 & \underline{71.3} / \underline{38.3} / 28.7 & 68.1 / 31.1 / 34.9 & -- & 71.0 / 33.9 / \underline{42.3} & 69.1 / 31.6 / 37.3 & \textbf{79.5} / \textbf{43.4} / \textbf{46.9} \\
\emph{Mean}   & 55.1 / 27.4 / 20.3 & 76.5 / \underline{49.3} / 50.5 & 69.2 / 33.5 / 36.4 & -- & 72.0 / 34.9 / 44.8 & \underline{78.2} / 41.1 / \underline{57.0} & \textbf{87.3} / \textbf{55.0} / \textbf{67.7} \\
\bottomrule
\end{tabular}
}
\end{table*}

\paragraph{Implementation Details.} We employ C-RADIOv4-h as RADIOlogy's backbone, a ViT-H/16 trunk
distilling SigLIP2-g, DINOv3, and SAM3 into one encoder. We encode every image once at $448{\times}448$ (a $28{\times}28$ patch grid) and evaluate pixel
maps after bilinear upsampling to $518{\times}518$ with a Gaussian blur ($\sigma{=}4$),
matching the resolution convention used by AdaCLIP~\cite{cao2024adaclip}.

Every numeric hyperparameter (context size $k{=}5$, SAM region counts $K \in \{8,32\}$, affinity refinement temperature $\tau=0.1$ and decay $\alpha=0.5$)
is identical across all fourteen benchmarks and is set to reasonable configurations given the input resolution and the number of object-parts in standard anomaly detection datasets. We provide a sensitivity analysis of all meaningful hyperparameters in the supplementary material (\cref{sec:sensitivity-analyses}), which demonstrates robustness over a reasonable range of values. Finally, the only domain-specific parameters are the state vocabulary and the two
foreground-gate thresholds of Section~\ref{sec:method-mask}, both fixed once by domain-partitioned leave-one-dataset-out calibration. While executed per target dataset, both thresholds collapse to the same value per domain: $0.016$ for the industrial and $0.014$ for the medical datasets.

\paragraph{Throughput Protocol.} Latency in Figure~\ref{fig:teaser-auroc-throughput} is
measured using MVTec-AD images and employing an NVIDIA H100 GPU, 
16 CPU cores, and 120 GB of RAM. For each method,
we batch inference at a size of 16 whenever its released code supports batching
(ours, CLIP, WinCLIP) and fall back to a batch size of 1 only where there is an architectural constraint that prevents it. This includes both MRAD-TF and FADE, which have retrieval algorithms implemented only for a batch size of 1, and SAA+, where the SAM segmentation pipeline prevents parallelization. Loading checkpoints, as well as memory-bank constructions, are excluded for fairness.

\subsection{Quantitative Results}

\Cref{tab:trainfree-image} contains the anomaly classification results for all methods and datasets included in this work (per-category results are in the supplementary material, \cref{sec:per-category-results}). Our method significantly outperforms all other baselines on this specific task by a large margin in both industrial and medical datasets across all metrics except AP, where CLIP~\cite{radford2021learning} in combination with our margin scoring rule slightly outperforms RADIOlogy in MPDD. The strength of the scoring rule in CLIP is further demonstrated in its average ranking, where it places second on all classification metrics. This highlights the benefit of following a contrastive approach to VLM-based anomaly detection irrespective of backbone: a language-aligned vision embedding containing an anomaly can only be scored properly with a normality baseline.

Additionally, the localization results in \cref{tab:trainfree-pixel} further demonstrate the superior performance of our method in the ZSAD training-free regime. While RADIOlogy does not obtain the best results across all datasets, it achieves the best AUROC results in five out of seven industrial datasets and the best performance in all medical datasets. Moreover, it has the best average performance in both domains, outperforming all other methods in the industrial domain by at least 3.5 points and by between 5.7 and 10.7 points in the medical datasets.

\subsection{Qualitative Results}

\begin{figure}
    \centering
    \includegraphics[width=1.0\linewidth]{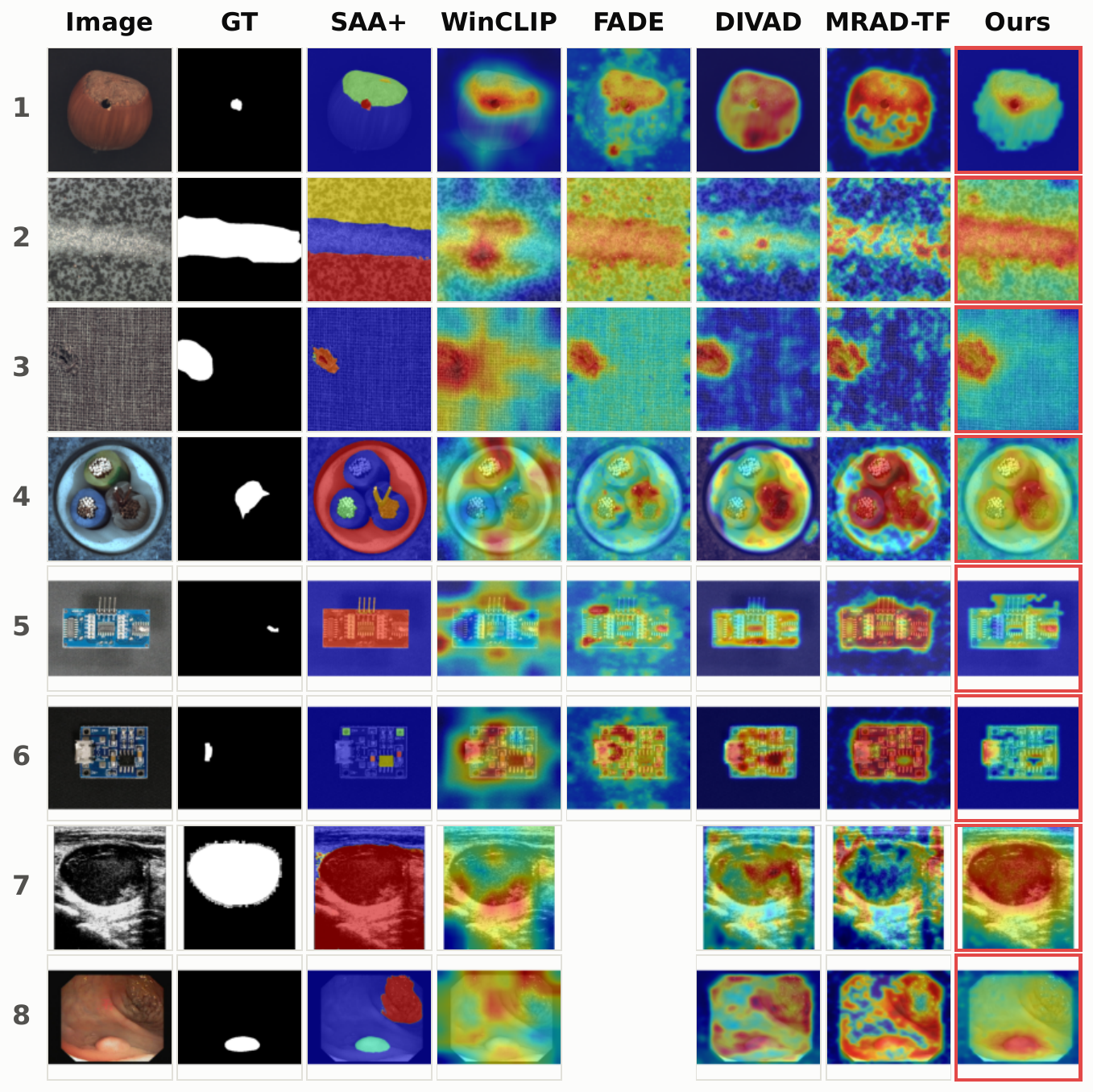}
    \caption{\textbf{Qualitative comparison.} Our method compared with the published baselines. All images are generated by resizing the outputs to a common resolution (518). Results from  industrial (MVTec and VisA) and medical datasets (TN3K and ClinicDB) are included. \textbf{GT} depicts the ground truth, and \textbf{Ours} the output of RADIOlogy. FADE is excluded from the medical datasets (\cref{sec:exp-setup}).}
    \label{fig:qualitative-results}
\end{figure}

\Cref{fig:qualitative-results} presents a comparison between the anomaly maps generated by RADIOlogy and the other five published baselines. We can observe that the anomaly regions generated by RADIOlogy are significantly better localized compared to the other methods, which often over-commit to over-sized anomaly regions (\eg MRAD and DIVAD) or have visible anomaly score peaks in non-anomalous locations (\eg SAA+, FADE, and WinCLIP). We attribute this localization improvement to the better suited SigLIP-distilled backbone and our anomaly segmentation refinements, which help reduce feature noise and increase object-part awareness. Moreover, since image-spanning objects (row 4) share more visual characteristics with textures than with object categories, the flexible mask language gate correctly avoids masking.

\subsection{Ablation Study}

\paragraph{Segmentation Refinement Components.}

\Cref{tab:component-ablation} portrays the effect of the anomaly segmentation refinement steps we employ to increase the quality of raw patch margin scores. The table shows that all components have a positive effect on segmentation performance, as hypothesized. Additionally, we can observe that the context pooling alone raises performance above the training-free SOTA, with an improvement of more than 4 points over the raw scores. This highlights the importance of contextualization in language-based anomaly detection: as discovered by WinCLIP, a single visual patch often does not contain enough information to differentiate an anomaly from typical texture variation, but a large enough pool of adjacent tokens contains sufficient semantics to be classified correctly relative to surface noise. Results describing the adequacy of each teacher head for each step are available in the supplementary material, \cref{sec:refinement-heads}.  

\begin{table}[t]
\centering
\caption{\textbf{Pixel-refinement component ablation.} Mean pixel AUROC and max-F1 over seven industrial benchmarks. The language-gated foreground prior
(Sec.~\ref{sec:method-mask}) is on in every row; only \textbf{context} pooling, \textbf{affinity}
refinement, and SAM3 object-part \textbf{region} pooling are toggled. Image-level metrics are
unaffected by all three components and are
therefore constant across rows, so they are omitted here. Bold marks the best row per column.}
\label{tab:component-ablation}
\begin{threeparttable}
\resizebox{0.75\linewidth}{!}{%
\begin{tabular}{ccc cc}
\toprule
Context & Affinity & Regions & AUROC & max-F1 \\
\midrule
            &            &            & 89.5 & 24.3 \\
\checkmark  &            &            & 93.7 & 28.9 \\
\checkmark  & \checkmark &            & 94.6 & 33.6 \\
\checkmark  & \checkmark & \checkmark & \textbf{95.2} & \textbf{37.8} \\
\bottomrule
\end{tabular}
}
\end{threeparttable}
\vspace{-10pt}
\end{table}

\paragraph{Multi-head Attention Pooling.}

To test our hypothesis that MAP-based encoders are fundamentally better suited for zero-shot anomaly segmentation, we measure the performance of standard \texttt{[CLS]} language-aligned backbones against encoders aligned with a MAP head in \cref{tab:maphead-ablation}. To isolate the effect of the backbone, we remove all pixel refinement components of our pipeline and segment the images with raw patch-level margin scores. The results show that MAP-based encoders substantially outperform the CLIP-based family. In fact, \texttt{[CLS]}-based dense features perform worse than random chance, implying they are anti-correlated with anomalies. In contrast, performance at the image-level behaves well in CLIP encoders, showcasing a correct correlation between anomaly prompts and global CLIP embeddings, as reported in the literature~\cite{jeong2023winclip, li2024fade}.

\begin{table}[t]
\centering
\caption{\textbf{Dense patch-text alignment by pooling architecture.} Anomaly maps are computed on the MVTec-AD dataset using the raw
per-patch margin without refinements. Bold marks the
best value per column, underline the second-best, and italics performance below chance. Alongside CLIP and SigLIP model variations, we include the performance of PE-Core~\cite{bolya2026perception}, whose attention-pooled summary embeddings are close to SigLIP's MAP head. The input is resized to the training resolution of the models when available, otherwise it is set to match the original SigLIP's native patch grid for MAP encoders, and CLIP (ViT-B/16)~\cite{radford2021learning} for CLIP variations.}
\label{tab:maphead-ablation}
\resizebox{\linewidth}{!}{%
\begin{tabular}{l l ccc}
\toprule
Encoder & Pooling & Pixel AUROC & Pixel max-F1 & Image AUROC \\
\midrule
SigLIP (so400m)~\cite{zhai2023sigmoid}         & MAP head & \underline{84.2} & \underline{29.3} & 91.7 \\
SigLIP2 (so400m, p14)~\cite{tschannen2025siglip}  & MAP head & 83.6 & 27.8 & 94.6 \\
SigLIP2-g ~\cite{tschannen2025siglip}  & MAP head & 81.1 & 27.3 & \textbf{96.4} \\
PE-Core (ViT-L/14)~\cite{bolya2026perception}          & MAP head & 75.2 & 26.1 & 91.6 \\
RADIOv2.5-h (SigLIP distilled)~\cite{heinrich2025radiov2} & MAP head distilled & \textbf{87.0} & \textbf{30.9} & 92.8 \\
C-RADIOv4-h (SigLIP2 distilled)~\cite{ranzinger2026c}      & MAP head distilled & 83.3 & 26.5 & \underline{95.3} \\
\midrule
CLIP (ViT-L/14)~\cite{radford2021learning}      & \texttt{[CLS]} + attn pool & \textit{34.7} & 6.4 & 88.2 \\
CLIP (ViT-B/16)~\cite{radford2021learning}          & \texttt{[CLS]} + attn pool & \textit{40.7} & 7.6 & 80.8 \\
WinCLIP's CLIP variant~\cite{jeong2023winclip}           & \texttt{[CLS]} + attn pool & \textit{20.6} & 6.2 & 90.4 \\
FADE's CLIP variant~\cite{li2024fade}                    & \texttt{[CLS]} + attn pool & \textit{44.3} & 7.5 & 87.8 \\
DFN5B-CLIP~\cite{fang2024dfnclip}                              & \texttt{[CLS]} + attn pool & \textit{18.9} & 6.3 & 92.7 \\
RADIOv2.5-h (DFN5B-CLIP distilled)~\cite{heinrich2025radiov2}                             & \texttt{[CLS]} + attn pool & \textit{17.7} & 6.4 & 93.1 \\
\bottomrule
\end{tabular}}
\end{table}

%% file: sec/5_conclusions.tex
\section{Conclusion}
\label{sec:conclusion}

In this work, we showed that the computational complexity of windowed execution is not a fundamental prerequisite for prompt-based, zero-shot visual anomaly detection. By systematically analyzing the token-pooling dynamics of vision-language architectures, we demonstrated that multi-head attention pooling inherently preserves fine-grained, patch-level text alignment out of the box, overcoming the structural token-mixing limitations that have historically degraded standard \texttt{[CLS]}-based CLIP encoders. Exploiting this property alongside the distilled representations exposed by the agglomerative C-RADIOv4 backbone, our unified RADIOlogy framework computes global anomaly margins in a single forward pass while applying training-free pixel refinements to produce sharp, localized defect segmentations. Across fourteen industrial and clinical benchmarks, our approach sets a new state of the art in the training-free regime, delivering substantial gains over prior baselines while operating at a fraction of their inference latency.

\paragraph{Limitations and Future Work.}
Our framework exhibits several limitations that present opportunities for future research. The heuristic foreground prior assumes nominal background borders, which fails for off-center objects that touch the image border. Incorporating depth estimation heads into agglomerative encoders could establish a robust geometric boundary to resolve this. Additionally, a performance gap persists between anomaly detection and localized segmentation. To tackle this, future directions include integrating multi-resolution token representations to capture fine structural defects and fine-tuning the C-RADIOv4 backbone to bridge the gap with trained ZSAD methods. 

\section*{Acknowledgements}
This work was carried out within the Xecs Marisens project.

%% file: sec/6_supplementary.tex
\section{Hyperparameter Sensitivity Analysis}
\label{sec:sensitivity-analyses}

\subsection{Context Pooling Window Size}
\label{sec:region-pooling-cluster-set}

\begin{figure}
    \centering
    \includegraphics[width=0.5\linewidth]{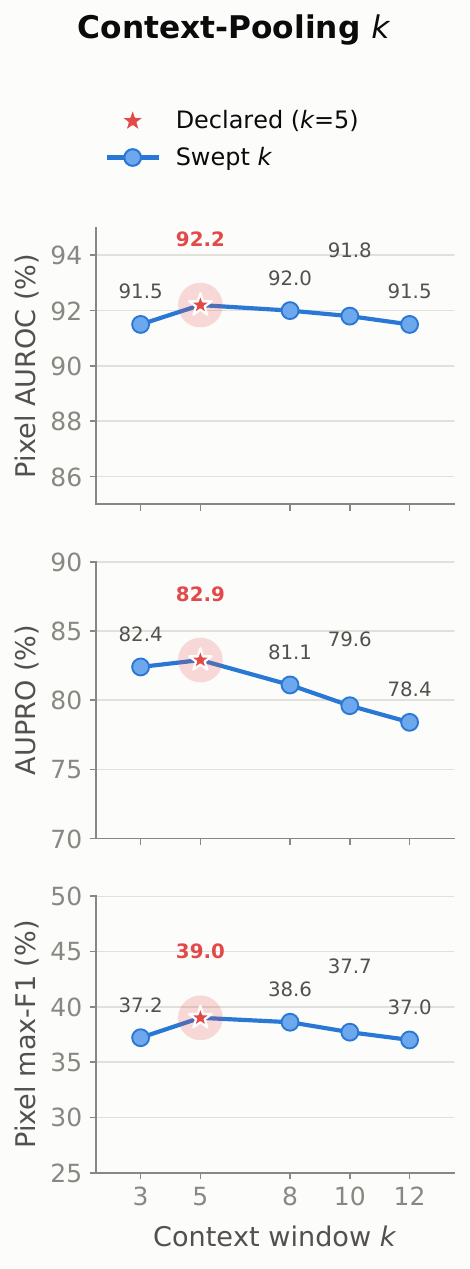}
    \caption{\textbf{Effect of the context pooling window size $k$.} Ablation of the effect of the window size in the context pooling mechanism over the raw patch margin scores. Measured on MVTec-AD~\cite{bergmann2019mvtec}. AUPRO is computed within a 0.3 limit.}
    \label{fig:ctxk-ablation}
\end{figure}

In \cref{fig:ctxk-ablation}, we can observe the performance of RADIOlogy as we vary the size of the context pooling window from 3 to 12 on the MVTec-AD~\cite{bergmann2019mvtec} dataset. All other hyper-parameters remain fixed. The results demonstrate robustness in both pixel AUROC and max-F1, the metrics reported in the main tables of this work. However, they do demonstrate sensitivity in the area under the per region overlap (AUPRO), larger windows have a negative effect compared to the reported hyper-parameters. This is to be expected, as excessively diffusing the margin scores results in less defined contours, an effect that is particularly detrimental for overlap-based metrics such as AUPRO. 

\subsection{Region Pooling Cluster Set}
\label{sec:context-pooling-window-size}

\begin{figure}
    \centering
    \includegraphics[width=0.5\linewidth]{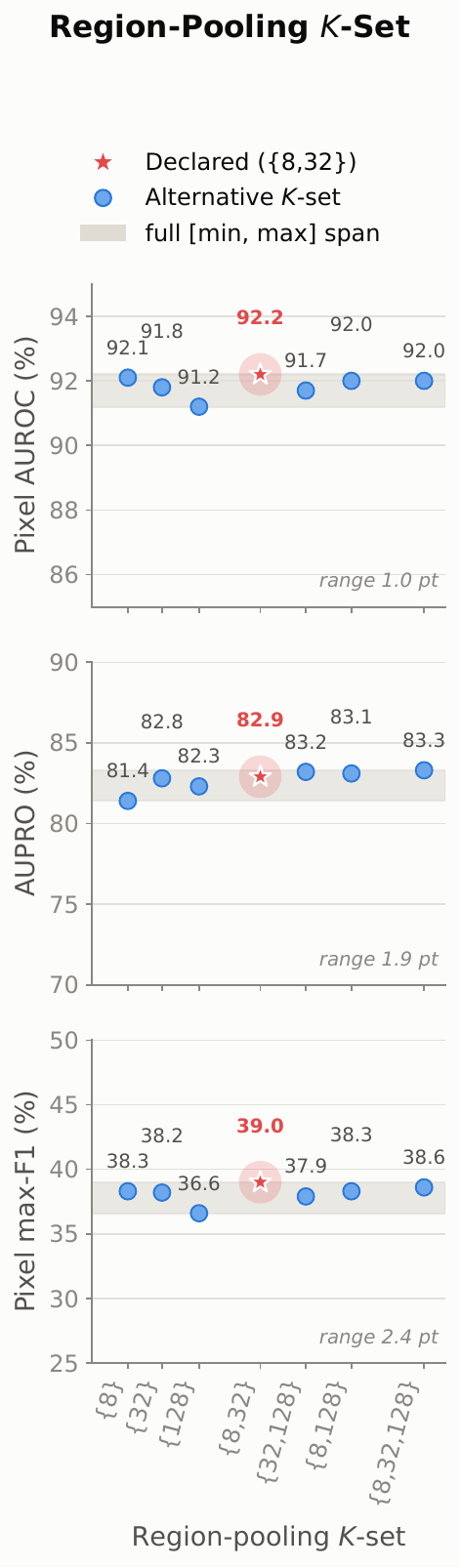}
    \caption{\textbf{Effect of the region pooling cluster set.} Ablation of the region pooling cluster set over the features produced by the SAM head of C-RADIOv4-h~\cite{ranzinger2026c}. All combinations over 8, 32 and 128 are provided. AUPRO is computed with a 0.3 limit.}
    \label{fig:regkset-stability}
\end{figure}

\Cref{fig:regkset-stability} showcases the effect of different $k$-means cluster-set configurations under our method in the MVTec-AD dataset. As our approach supports clustering with multiple cluster numbers, we vary the sets with three quantities: 8, 32, and 128. The results demonstrate a high degree of stability within multiple cluster combinations, with multi-cluster sets generally performing slightly better than single cluster results. We hypothesize that this advantage is derived from multi-cluster setups covering more edge cases and object configurations within the dataset, where single cluster values can be either too coarse or too fine-grained for some MVTec categories (\eg the hazelnut category requires only a few clusters, while the complexity of objects such as cables or transistors benefits from additional cluster capacity).     

\subsection{Affinity Refinement Steps}

\begin{figure}
    \centering
    \includegraphics[width=0.5\linewidth]{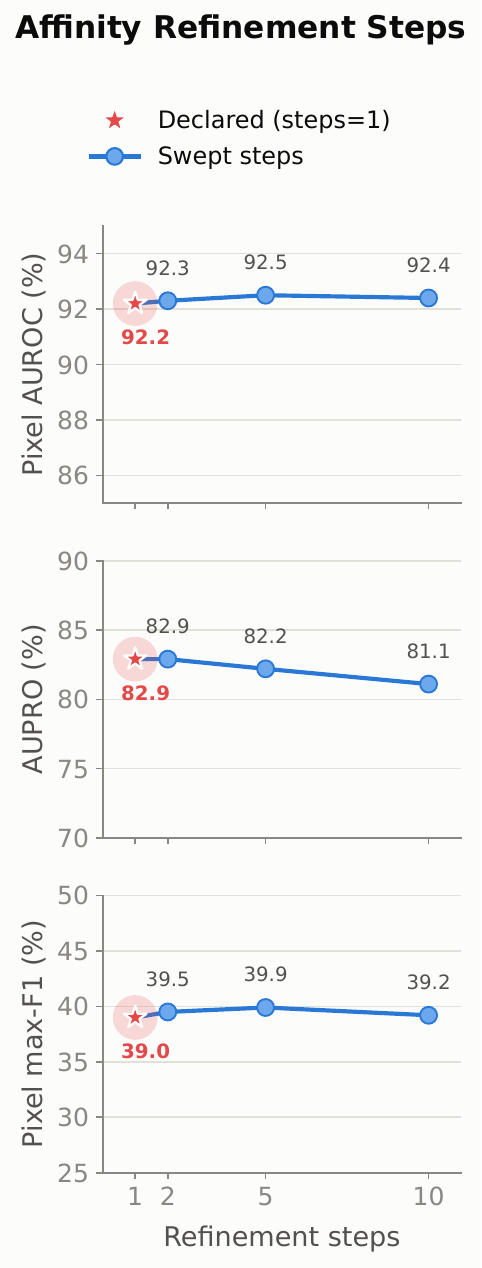}
    \caption{\textbf{Effect of the number of affinity refinement steps.} Ablation of the number of affinity refinement steps over the features produced by C-RADIOv4-h trunk encoder. The steps included range from 1 (declared configuration) to 10. AUPRO is computed with a 0.3 limit.}
    \label{fig:affinity-refinement-steps-ablation}
\end{figure}

A final sensitivity analysis is provided in \cref{fig:affinity-refinement-steps-ablation}. It illustrates performance stability within a moderate range of affinity steps. Increasing the number of steps leads to a slight improvement in pixel AUROC and max-F1. However, they tend to decrease AUPRO slightly, likely as a result of the dilution effect observed in the context window size ablation.

\section{Extended Ablation Study}

\subsection{Refinement Features Origin}
\label{sec:refinement-heads}

We perform an ablation study on the adequacy of the different features exposed by C-RADIOv4 in the segmentation refinement steps of \cref{sec:method-pixel}, depicted in \cref{tab:head-substrate-ablation}. The features analyzed include the C-RADIOv4 raw features and the embeddings obtained with the SAM3~\cite{carion2026sam}, DINOv3~\cite{simeoni2025dinov3}, and SigLIP2~\cite{tschannen2025siglip} heads. The results support our hypothesis that the SAM substrate is the most suitable space for object-part clustering: it obtains the best AUPRO and max-F1 (+0.4 and +1.9 points over the RADIO trunk, the next best substrate), while matching the RADIO trunk on pixel AUROC. For affinity refinement, the differences between substrates are smaller (at most 0.6 AUPRO and 1.1 max-F1), with the RADIO trunk performing best on all three metrics. This could be caused by RADIO's trunk features being of higher semantic quality compared to DINOv3, a reasonable result given its SigLIP2 supervision, which infuses it with a weakly supervised semantic signal that DINOv3 lacks. Finally, to demonstrate the need to combine different heads of RADIO, we include results for using the SigLIP2 head in the two steps individually and in both steps simultaneously. The SigLIP2 head is competitive for affinity refinement, but it is clearly worse than SAM for region pooling (2.2 points lower max-F1), and using it for both steps yields the lowest results in the table.

\begin{table}[t]
\centering
\caption{\textbf{Effect of different feature sets in the anomaly segmentation refinement steps.} Pixel-level AUROC (\%), AUPRO (\%, 0.3 limit) and max-F1 (\%), mean
over MVTec-AD, VisA~\cite{zou2022spot}, BTAD~\cite{mishra2021vt}, MPDD~\cite{jezek2021deep}, KolektorSDD (KSDD)~\cite{tabernik2020segmentation}, DAGM~\cite{wieler2007weakly} and DTD-Synthetic~\cite{cimpoi2014describing, aota2023zero}. The declared configuration
(Sec.~\ref{sec:method-pixel}) uses RADIO's trunk features for affinity diffusion
and its SAM3-adaptor features for region clustering. Each of the first two blocks varies the
features of one mechanism while the other is kept at its declared setting. Bold marks the best row per block and
column.}
\label{tab:head-substrate-ablation}
\begin{threeparttable}
\resizebox{\linewidth}{!}{%
\begin{tabular}{llccc}
\toprule
Mechanism & Substrate & AUROC & AUPRO & max-F1 \\
\midrule
\multirow{4}{*}{Affinity refinement}
  & DINO head             & 95.1 & 85.4 & 36.7 \\
  & RADIO trunk (declared) & \textbf{95.2} & \textbf{85.9} & \textbf{37.8} \\
  & SAM head              & 95.0 & 85.3 & 37.1 \\
  & SigLIP2 head          & 95.1 & 85.6 & 37.4 \\
\midrule
\multirow{4}{*}{Region pooling}
  & SAM head (declared)   & 95.2 & \textbf{85.9} & \textbf{37.8} \\
  & RADIO trunk           & \textbf{95.3} & 85.5 & 35.9 \\
  & DINO head             & 94.8 & 84.1 & 33.7 \\
  & SigLIP2 head          & 94.6 & 85.3 & 35.6 \\
\midrule
\multirow{2}{*}{Both mechanisms, jointly}
  & Declared (RADIO trunk aff., SAM reg.) & \textbf{95.2} & \textbf{85.9} & \textbf{37.8} \\
  & SigLIP2 head (both)            & 94.2 & 84.5 & 34.4 \\
\bottomrule
\end{tabular}
}
\end{threeparttable}
\end{table}

\subsection{Effect of the Foreground Mask}
\label{sec:effect-mask}

In order to demonstrate the benefits of foreground masking, we show the effect of removing the mask (No mask), employing the mask but removing its language gate (Mask, no gate), and the results for our declared configuration (Mask + language gate (declared)). We evaluate this effect on all the industrial datasets of our evaluation suite, as the benefit of the mask and gate is only apparent across domains. \Cref{tab:foreground-gate-mean} contains the result of this ablation and demonstrates the positive effect of both our simple foreground mask and the declared language-gate. Both components improve the overall anomaly segmentation performance significantly and complementarily. Masking alone results in a $\sim$1 point improvement in AUROC and max-F1, while combining both techniques improves pixel AUROC by 2.4 and 3.1 points in AUPRO over the unmasked pipeline.

\begin{table}
\centering
\caption{\textbf{Effect of the foreground mask and language gate on the seven industrial datasets.} Pixel-level AUROC (\%), AUPRO (\%, 0.3 limit) and max-F1 (\%) mean over the seven industrial datasets, isolating the foreground stage of Sec.~\ref{sec:method-mask} from
the declared pixel-refinement pipeline (context pooling,  RADIO affinity refinement, SAM3
region pooling), which is held fixed across all three rows. \emph{Mask, no gate} applies
the border-colour prior to every image unconditionally; \emph{Mask + language gate} adds
the object/surface abstention margin that decides per image whether to apply it (the
declared configuration).}
\label{tab:foreground-gate-mean}
\begin{threeparttable}
\resizebox{0.8\linewidth}{!}{
\begin{tabular}{lccc}
\toprule
Foreground stage & AUROC & AUPRO & max-F1 \\
\midrule
No mask                          & 92.8 & 82.8 & 37.0 \\
Mask, no gate                    & 93.7 & 84.2 & 37.8 \\
Mask + language gate (declared)  & \textbf{95.2} & \textbf{85.9} & \textbf{37.8} \\
\bottomrule
\end{tabular}
}
\end{threeparttable}
\end{table}

\subsection{Effect of the Medical Vocabulary}
\label{sec:medical-vocabulary}

To evaluate the effect of our medical state vocabulary, we replace the medical descriptors in the reported pipeline with the same vocabulary employed in the industrial benchmarks. We perform the same replacement for SAA+~\cite{cao2023segment} and WinCLIP~\cite{jeong2023winclip} to test whether the domain-adaptation vocabulary translates across techniques. The results can be observed in \cref{tab:medical-vocab-detection} and showcase a significant improvement over industrial descriptors for RADIOlogy. However, the effect is not conclusive for the other two techniques. While localization improves significantly in WinCLIP, detection is dramatically decreased by the medical prompts using the same method. Given that we did not observe this effect in our naive CLIP baseline, this hints at WinCLIP's anomaly detection score being overfit to its own industrial vocabulary, a reasonable prior given its industrial scope. Additionally, the medical prompts improve SAA+'s AUROC but do not consistently improve its other metrics, and the effect is much smaller than for RADIOlogy. This is explained by SAA+'s relatively low reliance on the textual priors, which are mostly employed to condition its text-prompted region proposals rather than to compute the anomaly score directly.

\begin{table}
\centering
\caption{\textbf{Medical anomaly-vocabulary ablation, detection.} Image-level AUROC~/~max-F1
(\%) on AdaCLIP's~\cite{cao2024adaclip} three detection-only medical benchmarks, industrial state words
(WinCLIP's own: \emph{damaged}, \emph{\{\} with
flaw/defect/damage}) vs.\ clinical state words (\emph{\{\} with a lesion/tumor/abnormality/
pathology}, \emph{abnormal \{\}}, \emph{diseased \{\}}), class names and every other
component held fixed within each method. For SAA+, the clinical state words, filled with each class name,
replace its general anomaly prompts and generic ``defect'' prompt; its property prompts are unchanged.
Bold marks the best vocabulary per method, per dataset, per metric.}
\label{tab:medical-vocab-detection}
\resizebox{\linewidth}{!}{%
\begin{tabular}{l cc cc cc}
\toprule
& \multicolumn{2}{c}{\textbf{Ours}} & \multicolumn{2}{c}{WinCLIP} & \multicolumn{2}{c}{SAA+} \\
\cmidrule(lr){2-3} \cmidrule(lr){4-5} \cmidrule(lr){6-7}
Dataset & Industrial & Clinical & Industrial & Clinical & Industrial & Clinical \\
\midrule
HeadCT~\cite{campos2018headct}   & 92.7 / 89.9 & \textbf{98.0} / \textbf{93.9} & \textbf{83.7} / \textbf{78.8} & 77.3 / 71.8 & 45.1 / \textbf{67.3} & \textbf{47.3} / 66.7 \\
BrainMRI~\cite{kanade2015brain} & 90.4 / 89.9 & \textbf{96.0} / \textbf{94.9} & \textbf{88.9} / \textbf{86.4} & 84.6 / 81.9 & 51.7 / \textbf{77.8} & \textbf{53.8} / 77.2 \\
Br35H~\cite{hamada2020br35h}    & 93.9 / 89.7 & \textbf{97.9} / \textbf{94.7} & \textbf{83.9} / \textbf{77.1} & 77.1 / 71.9 & 54.7 / \textbf{68.4} & \textbf{57.1} / 67.9 \\
\emph{Mean} & 92.3 / 89.8 & \textbf{97.3} / \textbf{94.5} & \textbf{85.5} / \textbf{80.8} & 79.7 / 75.2 & 50.5 / \textbf{71.2} & \textbf{52.7} / 70.6 \\
\bottomrule
\end{tabular}}
\end{table}

\begin{table}
\centering
\caption{\textbf{Medical anomaly-vocabulary ablation, localization.} Pixel-level
AUROC~/~max-F1 (\%) on AdaCLIP's four localization-only medical benchmarks selection, same
vocabulary pair and protocol as Table~\ref{tab:medical-vocab-detection}. The \emph{Ours,
Clinical} column is the declared configuration; the WinCLIP and SAA+ columns isolate the vocabulary swap only. Bold marks the best vocabulary per method, dataset, and metric.}
\label{tab:medical-vocab-localization}
\resizebox{\linewidth}{!}{%
\begin{tabular}{l cc cc cc}
\toprule
& \multicolumn{2}{c}{\textbf{Ours}} & \multicolumn{2}{c}{WinCLIP} & \multicolumn{2}{c}{SAA+} \\
\cmidrule(lr){2-3} \cmidrule(lr){4-5} \cmidrule(lr){6-7}
Dataset & Industrial & Clinical & Industrial & Clinical & Industrial & Clinical \\
\midrule
ISIC~\cite{gutman2016isic}     & 90.2 / \textbf{77.3} & \textbf{90.5} / 74.0 & \textbf{74.4} / \textbf{55.1} & 70.6 / 51.8 & 78.0 / 69.3 & \textbf{81.4} / \textbf{72.3} \\
ClinicDB~\cite{bernal2015wm} & 88.2 / 53.4 & \textbf{89.6} / \textbf{54.9} & 69.7 / 26.7 & \textbf{75.5} / \textbf{33.3} & 78.7 / 47.2 & \textbf{83.7} / \textbf{50.5} \\
ColonDB~\cite{tajbakhsh2015automated}  & 82.2 / 42.5 & \textbf{89.5} / \textbf{47.8} & 64.7 / 21.1 & \textbf{69.8} / \textbf{23.8} & \textbf{78.2} / \textbf{42.2} & 76.8 / 36.4 \\
TN3K~\cite{gong2023tn3k}     & 64.0 / 30.2 & \textbf{79.5} / \textbf{43.4} & 68.1 / \textbf{31.1} & \textbf{68.5} / 31.0 & \textbf{71.3} / \textbf{38.3} & 71.2 / 37.7 \\
\emph{Mean} & 81.1 / 50.8 & \textbf{87.3} / \textbf{55.0} & 69.2 / 33.5 & \textbf{71.1} / \textbf{35.0} & 76.5 / \textbf{49.3} & \textbf{78.3} / 49.2 \\
\bottomrule
\end{tabular}}
\end{table}

\subsection{Aspect Ratio Correction}
\label{sec:aspect-ratio-correction-ablation}

As mentioned in \cref{sec:exp-setup}, we observed that our pipeline is sensitive to extreme aspect-ratio deviations. The effect of this is most apparent in the KolektorSDD (KSDD) dataset, which contains many images formed by thin vertical strips with aspect ratios of more than 2.5:1. All other datasets have square-like resolutions hovering around the 1:1 ratio. Applying the tiling rule on KSDD results in a very significant performance gain in both detection and localization at the cost of two additional forward calls on smaller resolution images. We believe this aspect-ratio sensitivity is caused by two factors. First, KSDD's aspect ratio is extremely out-of-distribution for C-RADIOv4, which is trained on square images. Second, the square resizing of the input dramatically changes the geometry of the defects, diffusing the anomaly among the normal pixels. This diffusion causes a large number of anomalous pixels to be mislabeled, an effect apparent in the recall between tiling and not tiling the input: 50.19\% and 34.18\%, respectively. 

\begin{table}[t]
\centering
\caption{\textbf{The aspect-ratio band-tiling rule, KSDD.} Image- and pixel-level metrics
(\%) on KSDD, the dataset with the most severe aspect ratio deviation.
\emph{Without tiling} resizes each 500$\times$1260 frame directly to a square, an
anisotropic stretch; \emph{with tiling} (declared) splits it into 3 non-overlapping bands. Bold marks the best row per column.}
\label{tab:ksdd-aspect-ablation}
\begin{threeparttable}
\resizebox{\linewidth}{!}{%
\begin{tabular}{lccccc}
\toprule
& \multicolumn{2}{c}{Image} & \multicolumn{3}{c}{Pixel} \\
\cmidrule(lr){2-3} \cmidrule(lr){4-6}
Aspect handling & AUROC & max-F1 & AUROC & AUPRO & max-F1 \\
\midrule
Without correction & 93.5 & 70.7 & \textbf{98.5} & 89.9 & 25.9 \\
With correction (declared)         & \textbf{97.5} & \textbf{90.0} & \textbf{98.5} & \textbf{93.8} & \textbf{38.5} \\
\bottomrule
\end{tabular}
}
\end{threeparttable}
\end{table}

\subsection{Image Resolution}
\label{sec:image-resolution-ablation}

\begin{table}[t]
\centering
\caption{\textbf{Input resolution vs. image-level detection.} Mean image-level AUROC (\%) and max-F1 (\%) for the seven industrial datasets. Bold marks the best value per column.}
\label{tab:resolution-ablation-image}
\begin{threeparttable}
\resizebox{0.5\linewidth}{!}{
\begin{tabular}{lcc}
\toprule
Input resolution & AUROC & max-F1 \\
\midrule
224 & 89.6 & 86.6 \\
384 & 92.1 & 89.9 \\
448 (declared) & 92.6 & 90.5 \\
512 & \textbf{93.0} & 90.5 \\
672 & 92.7 & \textbf{90.8} \\
\bottomrule
\end{tabular}}
\end{threeparttable}
\end{table}

\begin{table}[t]
\centering
\caption{\textbf{Input resolution, per dataset and mean over seven industrial
benchmarks.} Pixel-level AUROC (\%), AUPRO (\%) and max-F1 (\%) for the declared full
pipeline run at five
encoder input resolutions. Evaluation
resolution is fixed at 518px; only the encoder's input resolution
and its resulting patch grid is modified. The context window $k$ scales with the
grid to maintain its physical receptive field roughly resolution-invariant
($k{=}3,5,5,7,9$); the region count $K\in\{8,32\}$ is held fixed. Bold marks
the best resolution per dataset per column.}
\label{tab:resolution-ablation-pixel}
\begin{threeparttable}
\resizebox{\linewidth}{!}{
\begin{tabular}{llccc}
\toprule
Dataset & Input resolution & AUROC & AUPRO & max-F1 \\
\midrule
\multirow{5}{*}{MVTec-AD}
  & 224 & 88.9 & 70.9 & 31.0 \\
  & 384 & 91.0 & 78.2 & 35.0 \\
  & 448 (declared) & \textbf{92.2} & \textbf{82.9} & \textbf{39.0} \\
  & 512 & 91.1 & 78.9 & 36.4 \\
  & 672 & 90.8 & 78.2 & 36.2 \\
\midrule
\multirow{5}{*}{VisA}
  & 224 & 90.4 & 67.5 & 14.9 \\
  & 384 & 93.6 & 80.3 & 18.1 \\
  & 448 (declared) & 93.7 & 83.1 & 20.4 \\
  & 512 & 94.2 & 82.8 & 19.0 \\
  & 672 & \textbf{94.5} & \textbf{83.9} & \textbf{20.6} \\
\midrule
\multirow{5}{*}{BTAD}
  & 224 & 91.4 & 62.3 & 31.3 \\
  & 384 & 93.7 & 71.5 & 39.8 \\
  & 448 (declared) & \textbf{96.2} & \textbf{77.8} & \textbf{52.9} \\
  & 512 & 93.3 & 71.5 & 41.7 \\
  & 672 & 92.4 & 70.9 & 42.1 \\
\midrule
\multirow{5}{*}{MPDD}
  & 224 & 92.4 & 76.9 & 18.2 \\
  & 384 & 93.8 & 83.4 & 22.9 \\
  & 448 (declared) & \textbf{94.4} & 85.5 & 24.7 \\
  & 512 & 94.2 & 84.7 & 25.1 \\
  & 672 & 94.1 & \textbf{85.7} & \textbf{26.5} \\
\midrule
\multirow{5}{*}{KSDD}
  & 224 & 95.9 & 85.7 & 27.3 \\
  & 384 & 96.8 & 90.0 & 36.6 \\
  & 448 (declared) & \textbf{98.5} & \textbf{93.8} & \textbf{38.5} \\
  & 512 & 98.4 & 92.0 & 36.8 \\
  & 672 & 98.5 & 92.8 & 37.7 \\
\midrule
\multirow{5}{*}{DAGM}
  & 224 & 74.2 & 54.3 & 16.9 \\
  & 384 & 89.7 & 78.6 & 36.0 \\
  & 448 (declared) & \textbf{95.3} & \textbf{88.6} & \textbf{43.7} \\
  & 512 & 89.3 & 76.6 & 34.2 \\
  & 672 & 89.1 & 75.8 & 33.6 \\
\midrule
\multirow{5}{*}{DTD-Synthetic}
  & 224 & 93.3 & 83.6 & 35.4 \\
  & 384 & 92.5 & 83.2 & 37.9 \\
  & 448 (declared) & \textbf{96.4} & \textbf{89.3} & \textbf{45.1} \\
  & 512 & 94.7 & 86.3 & 43.4 \\
  & 672 & 93.9 & 84.6 & 42.2 \\
\midrule
\multirow{5}{*}{Mean (7 benchmarks)}
  & 224 & 89.5 & 71.6 & 25.0 \\
  & 384 & 93.0 & 80.8 & 32.3 \\
  & 448 (declared) & \textbf{95.2} & \textbf{85.9} & \textbf{37.8} \\
  & 512 & 93.6 & 81.8 & 33.8 \\
  & 672 & 93.3 & 81.7 & 34.1 \\
\bottomrule
\end{tabular}
}
\end{threeparttable}
\vspace{-10pt}
\end{table}

The results of ablating the input resolution of RADIOlogy on all industrial datasets can be observed in \cref{tab:resolution-ablation-image} and \cref{tab:resolution-ablation-pixel}. They illustrate that anomaly classification results increase almost monotonically with resolution. This may be due to anomalies having a larger absolute footprint in the image (in terms of the number of pixels), which is reflected in the summary MAP representation. However, anomaly segmentation performance peaks non-monotonically at a resolution of 448. While higher resolutions increase performance in datasets with the highest resolution details, such as VisA and MPDD, performance decreases significantly in texture-centric datasets such as DAGM and DTD-Synthetic. We hypothesize that this is caused by a particular optimal point between the fine-grained details provided by larger resolutions and the dilution of information caused within each patch, since  they represent increasingly smaller regions of the input. 

\subsection{Effect of RADIO family and size}
\label{radio-variant-ablation}

To analyze the effect of encoder size and RADIO versions on RADIOlogy, we ablate the size and variant of RADIO in \cref{tab:radiov25-size-ablation}. The results demonstrate that, within the image level metrics, the newer C-RADIOv4 variants that distill newer state-of-the-art teachers~\cite{ranzinger2026c} obtain the best results. Moreover, performance remains consistent among different backbone sizes in the RADIOv2.5~\cite{heinrich2025radiov2} family, with metrics scaling slightly with backbone size. However, within the C-RADIOv4 family, performance drops substantially in the pixel metrics of the so400m model, falling below the results of any of the previous RADIOv2.5 variants. We hypothesize that this drop is attributable to SigLIP2's so400m architecture itself rather than model capacity: unlike the standard vision transformer variants underlying RADIOv2.5, so400m was obtained through a shape-optimized search targeting efficient global contrastive image-text alignment~\cite{alabdulmohsin2023getting}, which favors a compact, well-pooled embedding over the preservation of fine-grained spatial detail.

\begin{table}[t]
\centering
\caption{\textbf{Backbone family and size in the full RADIOlogy pipeline.} Different RADIO variants evaluated on MVTec-AD dataset using the RADIOlogy pipeline. Every cell is AUROC~/~max-F1 (\%). RADIOv2.5 uses its SigLIP
head (SigLIP v1 so400m teacher~\cite{zhai2023sigmoid}); its foreground-suppression gate threshold is
separately leave-one-dataset-out calibrated on a five-benchmark industrial
panel, since it operates in a different regime than C-RADIOv4's declared
value. C-RADIOv4 uses its declared SigLIP2-g head and gate. Bold
marks the best value per column, underline the second-best.}
\label{tab:radiov25-size-ablation}
\resizebox{\linewidth}{!}{%
\begin{tabular}{l cc}
\toprule
Encoder & Pixel & Image \\
\midrule
RADIOv2.5-B (siglip head) & \underline{90.36} / 37.78 & 91.02 / 92.56 \\
RADIOv2.5-L (siglip head) & 90.23 / \underline{39.63} & 93.86 / \underline{94.30} \\
RADIOv2.5-H (siglip head) & 90.21 / \textbf{40.10} & 93.36 / 93.08 \\
\midrule
C-RADIOv4-h (siglip2-g head)      & \textbf{92.15} / 38.98 & \textbf{96.00} / \textbf{95.14} \\
C-RADIOv4-so400m (siglip2-g head) & 86.34 / 36.97 & \underline{94.55} / 93.57 \\
\bottomrule
\end{tabular}
}
\end{table}

\subsection{Performance of the RADIO Teachers}

As our final ablation, we evaluate the effect of using the teachers of C-RADIOv4-h directly within the RADIOlogy pipeline. The goal of this ablation is to measure the quality of anomaly features lost due to RADIO's distillation process. The results are visible within \cref{tab:realteachers-image} and \cref{tab:realteachers-pixel}, containing anomaly detections and segmentation results, respectively, in the complete evaluation suite. They demonstrate that the difference between using C-RADIOv4-h and its much larger teacher ensemble is marginal in the industrial datasets but significant in the medical domain. Our hypothesis is that the distillation mechanism of RADIO, which compresses multiple model spaces into a single unified manifold, causes a detrimental effect on out-of-distribution generalization (\eg medical images).

\begin{table}
\centering
\caption{\textbf{Real teachers (SigLIP2-g and SAM3) vs. the distilled trunk in anomaly detection.} The table contains image-level AUROC~/~max-F1 (\%) for each dataset. Bold marks the best value per row.}
\label{tab:realteachers-image}
\resizebox{0.8\linewidth}{!}{%
\begin{tabular}{l cc}
\toprule
Dataset & C-RADIOv4-h (declared) & \textbf{Real teachers} \\
\midrule
\multicolumn{3}{l}{\textit{Industrial}} \\
\midrule
MVTec         & 96.0 / 95.1 & \textbf{96.3} / \textbf{95.6} \\
VisA          & \textbf{85.4} / \textbf{83.1} & 83.0 / 81.9 \\
BTAD          & \textbf{94.4} / 90.1 & 93.8 / \textbf{91.9} \\
MPDD          & \textbf{77.0} / \textbf{82.0} & 72.6 / 81.1 \\
KSDD          & 97.5 / \textbf{90.0} & \textbf{98.2} / 86.6 \\
DAGM          & \textbf{99.6} / \textbf{95.8} & 99.3 / 94.8 \\
DTD-Synthetic & \textbf{98.6} / \textbf{97.6} & 96.1 / 95.6 \\
\emph{Mean}   & \textbf{92.6} / \textbf{90.5} & 91.3 / 89.6 \\
\midrule
\multicolumn{3}{l}{\textit{Medical}} \\
\midrule
HeadCT        & 98.0 / 93.9 & \textbf{98.2} / \textbf{94.2} \\
BrainMRI      & 96.0 / 94.9 & \textbf{96.2} / \textbf{95.6} \\
Br35H         & 97.9 / 94.7 & \textbf{98.8} / \textbf{96.7} \\
\emph{Mean}   & 97.3 / 94.5 & \textbf{97.7} / \textbf{95.5} \\
\bottomrule
\end{tabular}
}
\end{table}

\begin{table}
\centering
\caption{\textbf{Real teachers (SigLIP2-g and SAM3) vs. the distilled trunk in anomaly segmentation.} Every cell is pixel-level AUROC~/~max-F1 (\%). Bold marks the best value per row.}
\label{tab:realteachers-pixel}
\resizebox{0.8\linewidth}{!}{%
\begin{tabular}{l cc}
\toprule
Dataset & C-RADIOv4-h (declared) & \textbf{Real teachers} \\
\midrule
\multicolumn{3}{l}{\textit{Industrial}} \\
\midrule
MVTec         & \textbf{92.2} / \textbf{39.0} & 88.3 / 38.9 \\
VisA          & 93.7 / 20.4 & \textbf{95.0} / \textbf{25.8} \\
BTAD          & \textbf{96.2} / \textbf{52.9} & 96.1 / 44.8 \\
MPDD          & \textbf{94.4} / 24.7 & 93.0 / \textbf{26.1} \\
KSDD          & 98.5 / 38.5 & \textbf{99.4} / \textbf{41.1} \\
DAGM          & 95.3 / 43.7 & \textbf{98.6} / \textbf{50.5} \\
DTD-Synthetic & \textbf{96.4} / 45.1 & 95.8 / \textbf{45.7} \\
\emph{Mean}   & \textbf{95.2} / 37.8 & 95.2 / \textbf{39.0} \\
\midrule
\multicolumn{3}{l}{\textit{Medical}} \\
\midrule
ISIC          & \textbf{90.5} / \textbf{74.0} & 88.0 / 70.2 \\
ClinicDB      & 89.6 / 54.9 & \textbf{92.2} / \textbf{65.2} \\
ColonDB       & \textbf{89.5} / 47.8 & 89.4 / \textbf{53.2} \\
TN3K          & 79.5 / 43.4 & \textbf{84.8} / \textbf{49.7} \\
\emph{Mean}   & 87.3 / 55.0 & \textbf{88.6} / \textbf{59.6} \\
\bottomrule
\end{tabular}
}
\end{table}


\begin{figure}[t]
    \centering
    \includegraphics[width=0.9\linewidth]{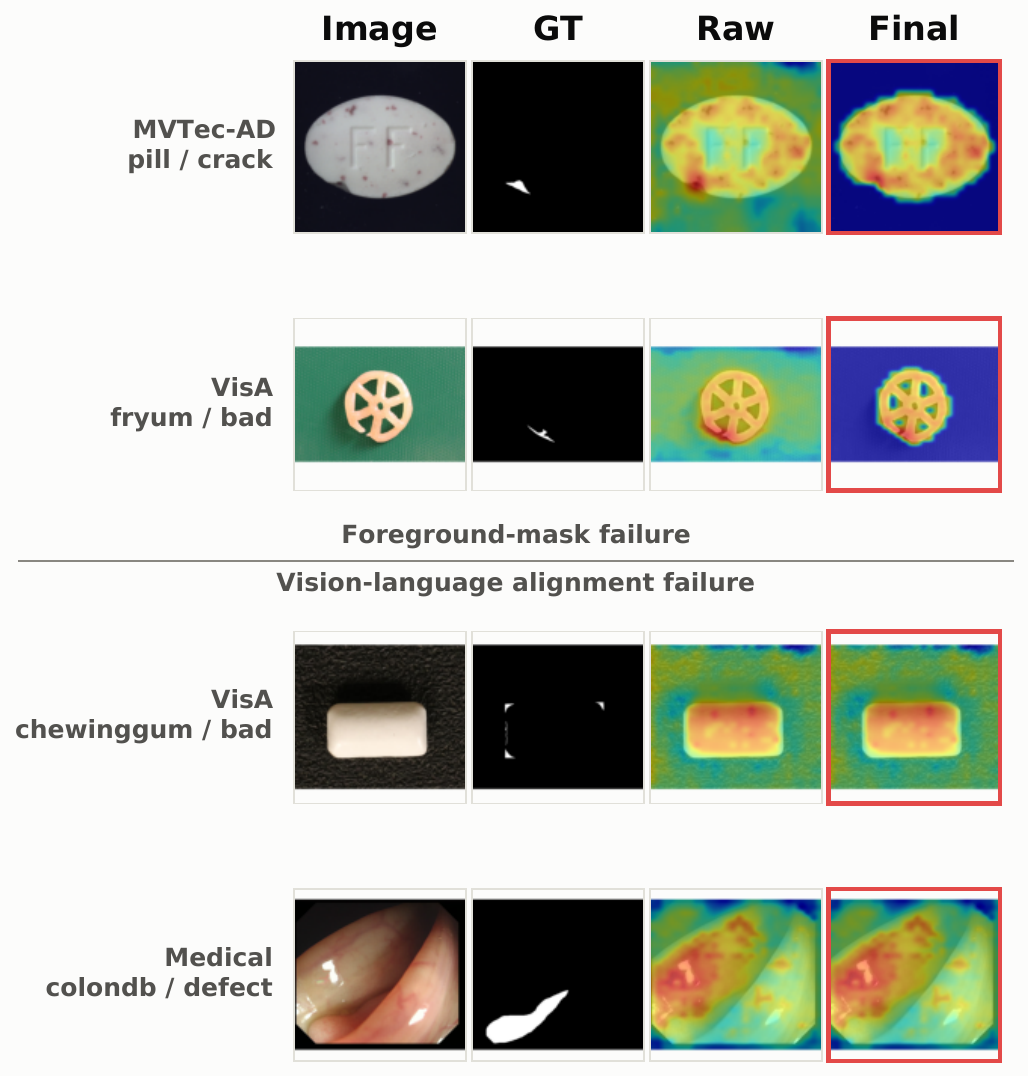}
    \caption{\textbf{Failure cases of RADIOlogy.} The figure depicts two classes of failure cases of our approach: masking errors and poor feature alignment.}
    \label{fig:failure-cases}
\end{figure}

\section{Failure Cases}
\label{sec:failure-cases}

\begin{table*}[t]
\centering
\caption{\textbf{Scoring prompt templates.} Photographic templates $\mathcal{T}$, paired by article. \texttt{\{\}}
is later filled with the state phrase (itself filled with the category
name).}
\label{tab:templates}

\resizebox{0.9\linewidth}{!}{
\begin{tabular}{@{}lll@{}}
\toprule
Style & \texttt{a \{\}} form & \texttt{the \{\}} form \\
\midrule
Cropped            & \texttt{a cropped photo of a \{\}.}   & \texttt{a cropped photo of the \{\}.} \\
Close-up           & \texttt{a close-up photo of a \{\}.}  & \texttt{a close-up photo of the \{\}.} \\
Bright             & \texttt{a bright photo of a \{\}.}    & \texttt{a bright photo of the \{\}.} \\
Dark               & \texttt{a dark photo of a \{\}.}      & \texttt{a dark photo of the \{\}.} \\
JPEG-corrupted     & \texttt{a jpeg corrupted photo of a \{\}.} & \texttt{a jpeg corrupted photo of the \{\}.} \\
Blurry             & \texttt{a blurry photo of a \{\}.}    & \texttt{a blurry photo of the \{\}.} \\
Plain              & \texttt{a photo of a \{\}.}           & \texttt{a photo of the \{\}.} \\
Small              & \texttt{a photo of a small \{\}.}     & \texttt{a photo of the small \{\}.} \\
Large              & \texttt{a photo of a large \{\}.}     & \texttt{a photo of the large \{\}.} \\
Visual inspection  & \texttt{a photo of a \{\} for visual inspection.} & \texttt{a photo of the \{\} for visual inspection.} \\
Anomaly detection  & \texttt{a photo of a \{\} for anomaly detection.} & \texttt{a photo of the \{\} for anomaly detection.} \\
\bottomrule
\end{tabular}}
\end{table*}

\begin{table*}[tp]
\centering
\caption{\textbf{Per-category results breakdown for RADIOlogy on the six multi-category industrial benchmarks.} The table contains image-level AUROC~/~max-F1~/~AP (\%) and pixel-level
AUROC~/~max-F1~/AUPRO (\%) under the declared configuration. Bold marks each
dataset's mean.}
\label{tab:percat-industrial}
\begin{minipage}[t]{0.48\linewidth}
\centering
\resizebox{\linewidth}{!}{%
\begin{tabular}{lcc}
\toprule
Category & Image AUROC / max-F1 / AP & Pixel AUROC / max-F1 / AUPRO \\
\midrule
\multicolumn{3}{l}{\textit{MVTec-AD}} \\
\midrule
Bottle      & 94.3 / 93.6 / 98.3 & 93.2 / 62.9 / 86.9 \\
Cable       & 92.7 / 89.5 / 95.8 & 88.2 / 27.3 / 73.2 \\
Capsule     & 92.4 / 93.4 / 98.5 & 96.6 / 35.7 / 89.4 \\
Carpet      & 100.0 / 100.0 / 100.0 & 98.9 / 57.9 / 93.4 \\
Grid        & 100.0 / 100.0 / 100.0 & 96.8 / 26.9 / 86.0 \\
Hazelnut    & 94.4 / 92.1 / 97.4 & 96.3 / 45.0 / 84.3 \\
Leather     & 99.9 / 99.5 / 100.0 & 97.8 / 25.2 / 93.9 \\
Metal Nut   & 94.7 / 94.7 / 98.8 & 75.9 / 35.4 / 69.3 \\
Pill        & 92.3 / 96.1 / 98.5 & 91.1 / 38.0 / 83.3 \\
Screw       & 98.5 / 97.1 / 99.5 & 96.0 / 13.2 / 84.1 \\
Tile        & 100.0 / 100.0 / 100.0 & 93.7 / 58.6 / 86.9 \\
Toothbrush  & 95.6 / 94.9 / 98.5 & 94.9 / 39.5 / 84.9 \\
Transistor  & 85.8 / 77.7 / 84.3 & 71.3 / 17.6 / 47.2 \\
Wood        & 100.0 / 100.0 / 100.0 & 96.4 / 58.6 / 93.5 \\
Zipper      & 99.5 / 98.7 / 99.9 & 95.2 / 42.9 / 87.1 \\
\midrule
\textbf{Mean} & \textbf{96.0 / 95.1 / 98.0} & \textbf{92.2 / 39.0 / 82.9} \\
\midrule
\multicolumn{3}{l}{\textit{VisA}} \\
\midrule
Candle      & 97.5 / 92.6 / 97.8 & 96.2 / 22.0 / 92.1 \\
Capsules    & 88.2 / 84.4 / 93.7 & 96.0 / 28.6 / 76.3 \\
Cashew      & 95.7 / 93.4 / 98.2 & 91.5 / 16.7 / 84.6 \\
Chewing Gum & 98.6 / 96.0 / 99.4 & 97.4 / 41.7 / 79.6 \\
Fryum       & 86.6 / 85.4 / 94.0 & 94.2 / 29.3 / 87.8 \\
Macaroni1   & 87.9 / 80.9 / 88.7 & 97.9 / 18.3 / 89.7 \\
Macaroni2   & 70.8 / 69.0 / 74.6 & 96.6 / 5.3 / 86.9 \\
PCB1        & 65.1 / 69.0 / 69.0 & 87.4 / 5.1 / 84.3 \\
PCB2        & 75.7 / 75.2 / 74.5 & 90.7 / 11.4 / 69.5 \\
PCB3        & 68.4 / 67.6 / 71.2 & 86.7 / 6.0 / 70.4 \\
PCB4        & 92.9 / 87.3 / 93.0 & 95.3 / 37.3 / 85.0 \\
Pipe Fryum  & 97.7 / 96.6 / 98.9 & 94.7 / 23.8 / 90.9 \\
\midrule
\textbf{Mean} & \textbf{85.4 / 83.1 / 87.8} & \textbf{93.7 / 20.4 / 83.1} \\
\midrule
\multicolumn{3}{l}{\textit{BTAD}} \\
\midrule
01 & 98.9 / 97.9 / 99.6 & 94.8 / 56.6 / 66.1 \\
02 & 87.8 / 94.3 / 98.1 & 95.0 / 55.9 / 69.4 \\
03 & 96.3 / 77.9 / 83.5 & 99.0 / 46.3 / 98.1 \\
\midrule
\textbf{Mean} & \textbf{94.4 / 90.1 / 93.7} & \textbf{96.2 / 52.9 / 77.8} \\
\bottomrule
\end{tabular}}
\end{minipage}%
\hfill
\begin{minipage}[t]{0.48\linewidth}
\centering
\resizebox{\linewidth}{!}{%
\begin{tabular}{lcc}
\toprule
Category & Image AUROC / max-F1 / AP & Pixel AUROC / max-F1 / AUPRO \\
\midrule
\multicolumn{3}{l}{\textit{MPDD}} \\
\midrule
Bracket Black & 78.8 / 81.1 / 81.8 & 96.5 / 8.8 / 90.1 \\
Bracket Brown & 57.5 / 79.7 / 71.5 & 88.8 / 6.4 / 80.3 \\
Bracket White & 55.9 / 72.3 / 56.0 & 97.2 / 1.2 / 92.1 \\
Connector     & 78.1 / 68.6 / 67.7 & 90.3 / 10.0 / 71.0 \\
Metal Plate   & 93.6 / 93.4 / 97.5 & 97.5 / 77.4 / 92.9 \\
Tubes         & 97.8 / 97.1 / 99.2 & 96.1 / 44.7 / 86.6 \\
\midrule
\textbf{Mean} & \textbf{77.0 / 82.0 / 78.9} & \textbf{94.4 / 24.7 / 85.5} \\
\midrule
\multicolumn{3}{l}{\textit{DAGM}} \\
\midrule
Class 1   & 97.1 / 81.6 / 88.8 & 86.4 / 29.8 / 72.0 \\
Class 2   & 100.0 / 100.0 / 100.0 & 97.9 / 37.9 / 96.4 \\
Class 3   & 100.0 / 100.0 / 100.0 & 96.7 / 52.3 / 94.7 \\
Class 4   & 99.4 / 89.1 / 96.0 & 92.1 / 6.8 / 76.5 \\
Class 5   & 100.0 / 100.0 / 100.0 & 98.1 / 73.5 / 95.6 \\
Class 6   & 100.0 / 100.0 / 100.0 & 98.6 / 77.3 / 97.2 \\
Class 7   & 99.7 / 94.5 / 98.4 & 87.5 / 17.1 / 63.9 \\
Class 8   & 99.9 / 97.0 / 99.6 & 96.5 / 52.6 / 92.1 \\
Class 9   & 99.8 / 96.4 / 97.4 & 99.6 / 45.6 / 98.8 \\
Class 10  & 100.0 / 99.0 / 99.8 & 99.3 / 44.4 / 98.5 \\
\midrule
\textbf{Mean} & \textbf{99.6 / 95.8 / 98.0} & \textbf{95.3 / 43.7 / 88.6} \\
\midrule
\multicolumn{3}{l}{\textit{DTD-Synthetic}} \\
\midrule
Blotchy 099     & 98.7 / 97.5 / 99.7 & 96.0 / 45.0 / 84.7 \\
Fibrous 183     & 99.6 / 99.4 / 99.9 & 98.8 / 53.6 / 95.8 \\
Marbled 078     & 97.4 / 96.9 / 99.4 & 92.8 / 26.4 / 81.2 \\
Matted 069      & 99.9 / 99.4 / 100.0 & 99.2 / 56.1 / 83.8 \\
Mesh 114        & 93.6 / 90.5 / 97.5 & 95.9 / 42.6 / 84.9 \\
Perforated 037  & 100.0 / 100.0 / 100.0 & 97.4 / 38.3 / 94.7 \\
Stratified 154  & 99.4 / 98.7 / 99.9 & 99.6 / 66.0 / 97.6 \\
Woven 001       & 99.6 / 98.6 / 99.9 & 92.1 / 34.9 / 81.0 \\
Woven 068       & 98.8 / 96.9 / 99.4 & 98.3 / 47.4 / 94.3 \\
Woven 104       & 98.9 / 97.5 / 99.8 & 95.9 / 46.5 / 92.0 \\
Woven 125       & 98.5 / 97.5 / 99.7 & 96.2 / 38.7 / 90.0 \\
Woven 127       & 99.1 / 98.1 / 99.3 & 94.3 / 46.0 / 92.0 \\
\midrule
\textbf{Mean} & \textbf{98.6 / 97.6 / 99.5} & \textbf{96.4 / 45.1 / 89.3} \\
\bottomrule
\end{tabular}}
\end{minipage}
\end{table*}

We illustrate failure cases of our RADIOlogy pipeline in \cref{fig:failure-cases} to demonstrate which elements should be targeted for improvement in future work. The first element is a failure of the masking strategy, which, in some situations, incorrectly masks the defect to be segmented; this failure mode is particularly common at object edges. A second failure mode is caused by the patch margin scores themselves, which can fail to detect anomalies entirely. The error in the VisA chewing gum category is likely caused by the lack of normal examples, which are required to understand the exact geometry of the object. The error in the last row (ColonDB) probably originates from poor vision-language alignment, which aligns superficial qualities such as light reflections and color variations with pathological properties. Replacing SigLIP with a medical vision-language encoder (\eg MedSigLIP~\cite{sellergren2026medgemmatechnicalreport}) could improve this misalignment.

\section{Per-Category Results}
\label{sec:per-category-results}

\Cref{tab:percat-industrial} contains per-category results of RADIOlogy for all the datasets that include multiple categories. Since all the medical datasets contain a single category, the exhibited datasets consist only of industrial benchmarks. They demonstrate generally consistent results among a variety of categories and object types. However, there are outlier categories that disproportionately decrease the average metrics, notably at the image level. These include categories such as PCB1 and multiple categories of MPDD, which are generally characterized by having large object surface variation (PCB) or large geometric variations (MPDD). We believe such characteristics are difficult to summarize through the relatively simple prompt ensemble we employed and would generally require more targeted and detailed descriptions.

\section{Prompt Ensemble Details}
\label{sec:prompt-ensemble-details}

The prompt ensemble used to score images is composed of two parts: a template and a state. Prompts are generated by substituting each state into the templates. The templates, visible in \cref{tab:templates}, are shared by industrial and medical evaluations and are obtained from WinCLIP. The states vary by domain and are present in \cref{tab:states}. For the gate used to suppress the foreground mask in texture categories, we replace the states in \cref{tab:states} with the states in \cref{tab:gate-vocab}.

\begin{table}[t]
\centering
\caption{\textbf{Scoring state vocabularies.} State vocabularies $\mathcal{S}$, by domain. \texttt{\{\}} is filled
with the category name (e.g.\ \texttt{bottle}, \texttt{brain MRI}).}
\label{tab:states}
\begin{subtable}{\linewidth}
\centering
\caption{Industrial (default)}
\resizebox{0.7\linewidth}{!}{
\begin{tabular}{@{}ll@{}}
\toprule
Normal states $\mathcal{S}_n$ & Anomalous states $\mathcal{S}_a$ \\
\midrule
\texttt{\{\}}                  & \texttt{damaged \{\}} \\
\texttt{flawless \{\}}         & \texttt{\{\} with flaw} \\
\texttt{perfect \{\}}          & \texttt{\{\} with defect} \\
\texttt{unblemished \{\}}      & \texttt{\{\} with damage} \\
\texttt{\{\} without flaw}     & \\
\texttt{\{\} without defect}   & \\
\texttt{\{\} without damage}   & \\
\bottomrule
\end{tabular}}
\end{subtable}

\vspace{1em}
\begin{subtable}{\linewidth}
\centering
\caption{Clinical (medical benchmarks)}
\resizebox{\linewidth}{!}{
\begin{tabular}{@{}ll@{}}
\toprule
Normal states $\mathcal{S}_n$ & Anomalous states $\mathcal{S}_a$ \\
\midrule
\texttt{\{\}}                     & \texttt{\{\} with a lesion} \\
\texttt{healthy \{\}}             & \texttt{\{\} with a tumor} \\
\texttt{normal \{\}}              & \texttt{\{\} with an abnormality} \\
\texttt{\{\} without lesion}      & \texttt{\{\} with pathology} \\
\texttt{\{\} without abnormality} & \texttt{abnormal \{\}} \\
\texttt{\{\} without pathology}   & \texttt{diseased \{\}} \\
\bottomrule
\end{tabular}
}
\end{subtable}
\end{table}

\begin{table}[t]
\centering
\caption{\textbf{Foreground/background gate vocabularies.} The abstention
gate that decides when to skip the foreground mask re-purposes the scoring
templates $\mathcal{T}$ (Table~\ref{tab:templates}) with 
foreground/background descriptors. $\theta$ refers to the gate's foreground threshold value (different on each domain).}
\label{tab:gate-vocab}
\begin{subtable}{\linewidth}
\centering
\caption{Industrial (default, $\theta = 0.016$)}
\resizebox{0.7\linewidth}{!}{
\begin{tabular}{@{}ll@{}}
\toprule
Foreground words $\mathcal{F}$ & Background words $\mathcal{B}$ \\
\midrule
\texttt{object} & \texttt{surface} \\
\texttt{thing}  & \texttt{texture} \\
\bottomrule
\end{tabular}}
\end{subtable}

\vspace{1em}
\begin{subtable}{\linewidth}
\centering
\caption{Clinical (medical benchmarks, $\theta = 0.014$)}
\resizebox{0.7\linewidth}{!}{
\begin{tabular}{@{}ll@{}}
\toprule
Foreground words $\mathcal{F}$ & Background words $\mathcal{B}$ \\
\midrule
\texttt{organ}   & \texttt{tissue} \\
\texttt{anatomy} & \texttt{texture} \\
\bottomrule
\end{tabular}}
\end{subtable}
\vspace{-10pt}
\end{table}

\section{Baseline Reproduction Details}
\label{sec:baseline-reproduction}

To ensure a fair comparison, every reproduced baseline is run from the authors' own released code at its published zero-shot configuration and scored with its own evaluation code. Before trusting a method on benchmarks its paper does not report, we first validate it on the benchmarks it does. Extending a method to new benchmarks only requires adding a data loader and the category names used in the prompts; neither the model nor the scoring is modified. Below we list every deviation from the released code and the reason for it, so that the reported numbers can be traced back to the original implementations.

\paragraph{WinCLIP.} We use the public \texttt{Accurate-WinCLIP-pytorch} reimplementation~\cite{zhou2024anomalyclip} with its original encoder (OpenCLIP ViT-B/16+, LAION-400M, 240px), window sizes and prompt ensemble, without Gaussian smoothing of the anomaly maps, as in the original method. Run unmodified, it reproduces that repository's published MVTec and VisA results exactly. For the remaining benchmarks, we only add a data loader and use the raw dataset folder names as class names, which reproduces the WinCLIP numbers cited by AdaCLIP on MVTec, VisA, BTAD, MPDD and DTD-Synthetic.

\paragraph{SAA+.} We run the full SAA+~\cite{cao2023segment} model, unmodified, from the authors' \texttt{SAA-plus} codebase: the class-specific language prompts, the object property prompts (area and location filters), the visual saliency rescoring, and the confidence prompting that merges the proposed masks into a continuous anomaly map. On MVTec-AD and VisA, we use the authors' own prompt files unmodified. For the remaining benchmarks, which have no released configuration, we reuse the MVTec property-prompt defaults and SAA+'s general anomaly prompts, extended with a generic ``defect'' prompt and filled with each category's name, the same class names used by every method, without tuning them per dataset. The medical benchmarks use the same prompts; replacing them with our medical vocabulary is evaluated in \cref{sec:medical-vocabulary}. Following SAA+'s evaluation script, maps and masks are evaluated at $400{\times}400$, normal images enter the pixel-level metrics with all-zero masks, and the image score is the maximum of the anomaly map. On MVTec-AD and VisA, our run reaches 37.7 and 33.0 pixel max-F1, against the 39.4 and 27.1 reported by the authors. The released code does not reproduce the published numbers: an independent run of it reports 37.65 on MVTec-AD, which the authors attribute to differences between code versions.\footnote{\url{https://github.com/caoyunkang/Segment-Any-Anomaly/issues/30}}

\paragraph{FADE.} FADE~\cite{li2024fade} is run without modification at its zero-shot configuration. Our run reads 90.1/89.2 and 74.7/91.1 image/pixel AUROC on MVTec and VisA against the published 90.0/89.6 and 75.6/91.5. We attribute this residual gap of at most 0.9 points to the unpinned software environment of the released code. Since FADE's state prompts are WinCLIP's industrial vocabulary, we do not evaluate it on the medical benchmarks.

\paragraph{MRAD-TF.} MRAD-TF~\cite{xu2026mrad} fuses the top-1\% mean of the anomaly map with a memory-retrieval score as $(1-k)\,s_\text{map} + k\,s_\text{mem}$. The paper describes an unweighted sum ($k{=}0.5$), whereas the released script defaults to $k{=}0.7$. With $k{=}0.5$, we reproduce the published image-level results on MVTec, BTAD, MPDD and DTD-Synthetic, while VisA is only reproduced with the script default $k{=}0.7$. We therefore report VisA with $k{=}0.7$ and every other benchmark with $k{=}0.5$. Pixel-level results do not depend on $k$. Following the authors, the retrieval memory is built from VisA, or from MVTec when VisA is the target. On DAGM, the gap to the published numbers stems from the test split: the authors evaluate on a balanced subset of 2000 normal and 2000 defective images, while we use the official test set (8050 images, 1054 defective) for every method.

\paragraph{DIVAD.} DIVAD~\cite{hicsonmez2026training} required four changes to its released code.
\begin{enumerate}
    \item The Stable Diffusion~2.1 checkpoints used for DDIM inversion are no longer hosted by Stability AI. We load them from a community mirror whose weights we verified to be byte-identical (SHA-256) to the original release.
    \item The released command passes \texttt{--use\_dino}, which is declared as \texttt{store\_false} and therefore \emph{disables} the DINO segmenter, falling back to a ResNet-50. We keep DINO enabled, as intended by the method.
    \item The list of MVTec texture categories exempted from the CutLER foreground mask contains \texttt{zipper} instead of \texttt{grid}. We correct it, so that masks are applied to object categories only, as described in the paper. With these first three changes, MVTec reads 75.9/88.1 image/pixel AUROC against the published 75.9/88.0.
    \item On BTAD, the released evaluation retrieves each ground-truth mask from the first three characters of the four-digit image name prefixed with a zero, so test image \texttt{0049} is paired with the mask of image \texttt{0004}. Apart from the first image of each category, every defective image is therefore scored against the mask of another image. Because the image labels are also derived from these masks, the mismatch affects detection as well as localization.
\end{enumerate}
As shown in \cref{tab:divad-btad}, restoring the released lookup reproduces the published BTAD results exactly, which confirms that the lookup explains the discrepancy. Pairing each image with its own mask raises DIVAD's results substantially, most notably the pixel AUROC, from 68.4 to 94.3. We report the corrected numbers in the main paper. Both BTAD runs follow the released protocol, without foreground masks.

\begin{table}[t]
\centering
\caption{\textbf{DIVAD on BTAD.} Image AUROC~/~max-F1~/~AP and pixel AUROC~/~max-F1~/~AUPRO (\%). The released ground-truth lookup pairs most test images with the mask of another image; correcting it yields the numbers reported in the main paper.}
\label{tab:divad-btad}
\resizebox{\linewidth}{!}{
\begin{tabular}{@{}lcc@{}}
\toprule
 & Image & Pixel \\
\midrule
Published                 & 79.8 / -- / --      & 68.4 / -- / --      \\
Released lookup    & 79.8 / 83.6 / 87.0  & 68.4 / 19.1 / 29.3  \\
Corrected lookup (ours)   & 82.4 / 88.0 / 91.3  & 94.3 / 52.5 / 69.9  \\
\bottomrule
\end{tabular}}
\end{table}